\documentclass{article}

\usepackage[preprint]{neurips_2026}

\usepackage[utf8]{inputenc}
\usepackage[T1]{fontenc}
\usepackage{microtype}
\usepackage{graphicx}
\usepackage{subcaption}
\usepackage{booktabs}
\usepackage{multirow}
\usepackage{placeins}
\usepackage{hyperref}
\usepackage{url}

\usepackage{amsmath}
\usepackage{amssymb}
\usepackage{mathtools}
\usepackage{amsthm}
\usepackage{siunitx}
\usepackage{tcolorbox}
\usepackage{enumitem}
\usepackage{algorithm}
\usepackage{algorithmic}

\usepackage[capitalize,noabbrev]{cleveref}

\theoremstyle{plain}

\theoremstyle{definition}

\theoremstyle{remark}

\usepackage[disable,textsize=tiny]{todonotes}

\title{Neural Slack Variables for Shape Constraints}

\author{%
  Ruben Wiedemann \\
  Imperial College London \\
  London, UK \\
  \texttt{r.wiedemann22@imperial.ac.uk} \\
  \And
  Antoine Jacquier \\
  Imperial College London \\
  London, UK \\
  \texttt{a.jacquier@imperial.ac.uk} \\
  \And
  Lukas Gonon \\
  University of St.\ Gallen \\
  St.\ Gallen, Switzerland \\
  \texttt{lukas.gonon@unisg.ch} \\
}

\begin{document}

\maketitle

\begin{abstract}
  Enforcing functional inequality constraints such as monotonicity and convexity in neural networks is a fundamental challenge in many industrial and scientific applications.
  Classical one-sided penalty methods, along with primal-dual methods gated by complementary slackness, provide constraint gradients only at violated locations, resulting in fragile satisfaction.
  Architectures that guarantee feasibility by construction, on the other hand, remain largely limited to elementary cases and impose additional inductive biases.
  We introduce \emph{neural slack variables}, a deep learning native primal-side approach that converts constraint enforcement into a regression problem by coupling the primary network with a jointly learned auxiliary network.
  The auxiliary network serves as a valid target for the primary network's constraint quantities, inducing feasibility and regularity.
  Neural slack variables achieve zero measured violations on dense-grid monotonicity and convexity test cases, where penalty and primal-dual baselines leave residual violations, and enable arbitrage-free learning of volatility surfaces, an open industrial challenge in quantitative finance.
\end{abstract}

\section{Introduction}
\label{sec:introduction}

In scientific computing and industrial domains such as control systems and quantitative finance, neural networks serve as learned function approximators on low-dimensional continuous domains.
The learned map must both reproduce data with high fidelity and respect known structural laws such as monotonicity, convexity, stability, or absence of arbitrage.
These are \emph{shape constraints}: conditions on the function and its derivatives that must hold everywhere on the domain, expressed as functional inequalities.
Standard neural architectures offer no intrinsic mechanism to preserve them, and unconstrained training routinely produces models that fit the data while remaining operationally inconsistent.

Specialized architectures that guarantee feasibility by construction (\emph{architectural constraints}) exist for elementary constraint types: Constrained Monotonic Neural Networks (CMNN) \cite{runjeConstrainedMonotonicNeural2023,sartorAdvancingConstrainedMonotonic2025} for monotonicity, Input-Convex Neural Networks (ICNN) \cite{amosInputConvexNeural2017} for convexity.
These architectures restrict the hypothesis class in ways that can make them stiffer to train and bias the fit toward low-frequency solutions \citep{sivaprasadCuriousCaseConvex2021}.

For general constraints and architectures, constraints must be enforced through the training procedure instead (\emph{soft constraints}).
The default method in deep learning is the \emph{penalty method}: a violation cost term is added to the objective, computed as an aggregate of the constraint residual; its weight determines how strongly constraint satisfaction is enforced relative to data fit.
\citet{ramirezPositionAdoptConstraints2025} advocate primal-dual methods as the principled alternative, which introduce Lagrangian multipliers that adaptively balance fit and feasibility. 
In the deep learning regime, however, both methods exhibit the same ``constraint drifting'' failure mode in practice:
For the penalty method, the violation term supplies no gradient wherever the network is already feasible.
Subsequent training steps can then reintroduce violations, which are corrected only after they appear, leading to repeated drift rather than stable satisfaction.
Primal-dual methods inherit constraint drifting for analogous structural reasons: under complementary slackness, the multiplier is zero in feasible regions, again eliminating the constraint gradient (\Cref{sec:background-lagrangian}).
The outcome for both methods is often spurious violations in trained models, which is a real practical challenge.
In control systems, a learned barrier certificate carries no guarantee unless it is exact (formal verification is an open research direction \citep{daiCounterexampleGuidedSynthesis2020,daiLyapunovstableNeuralnetworkControl2021,edwardsFossil20Formal2024}).
In quantitative finance, an implied volatility surface that admits even slight arbitrage is a significant issue for any downstream pricing and risk model \citep{gatheralArbitragefreeSVIVolatility2014,deschatresConvexVolatilityInterpolation2024,buehlerSANOSSmoothStrictly2026}. Ensuring fully satisfied constraints is a high-impact industrial challenge even for low-dimensional parametric approaches.
 
\begin{figure}[!t]
    \centering
    \includegraphics[width=\columnwidth]{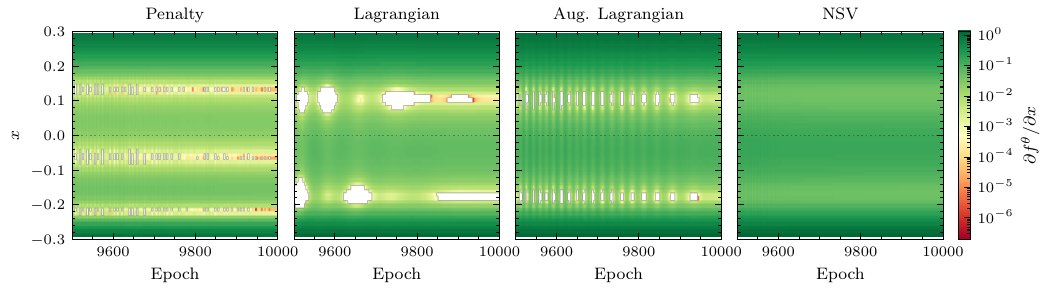}
    \caption{
        $\partial_x f^\theta(x)$ over last 500 training epochs for monotonicity test case in \Cref{sec:exp-monotone}.
        White (empty) regions indicate violations (negative derivatives), feasible values are colormapped.
        Penalty and Lagrangian methods exhibit ``constraint drift''; neural slack variables maintain stable constraint profiles.
      }
    \label{fig:drift-monotone}
\end{figure}

To overcome the issue of vanishing constraint gradients observed in the penalty method and primal-dual methods, we propose to tie the primary network's constraint profile to a jointly learned valid target, rather than penalizing only its violations.
Writing $f^\theta$ for the primary network, let $c_\theta := \mathcal{C}[f^\theta] \colon \Omega \to \mathbb{R}^m$ denote the constrained quantities.
We introduce an auxiliary neural network $s^\phi$---a learned analogue of the classical slack variable in constrained optimization---whose outputs are non-negative by construction (through an appropriate output activation function) and provide the valid target in constraint space.
The two networks are trained jointly, enforcing the slack-variable form
\begin{equation}\label{eq:slack-variables}
\mathcal{C}[f^\theta](x) - s^\phi(x) = 0, \qquad s^\phi(x) \geq 0,
\end{equation}
softly through a quadratic matching loss in constraint space.
The matching term couples the two networks: $s^\phi$ adapts to the data-driven shape of $c_\theta$, while $c_\theta$ is steered toward the valid target $s^\phi$.
As the matching residual shrinks, $c_\theta$ inherits the positivity of $s^\phi$, making $f^\theta$ feasible.
Crucially, because $s^\phi$ is a learned, finite-capacity approximation, the matching residual generally does not fully vanish, which leaves a constraint-space gradient after violations disappear and prevents the constraint profile from drifting back into violation.

We term $s^\phi$ a \emph{neural slack variable}.
In our experiments, neural slack variables overcome the drifting failure mode that we observe for penalty and primal-dual methods (\Cref{fig:drift-monotone}).
We identify an additional mechanism of the proposed method:
the neural slack loss transfers the regularity of $s^\phi$ to the constraint profile $c_\theta$ (\Cref{fig:monotone-aux-derivatives}), making the architecture of $s^\phi$ an explicit, controllable inductive bias on that profile.
We use this lever to constrain spectrally expressive primary networks (including SIRENs~\cite{sitzmannImplicitNeuralRepresentations2020} and Fourier features~\cite{tancikFourierFeaturesLet2020}); in fixed-grid settings, this can help keep satisfaction stable between collocation points.

Viewed structurally, neural slack variables decompose the feasibility problem at the constraint operator.
Once the profile $c_\theta = \mathcal{C}[f^\theta]$ is formed, feasibility is simply its pointwise non-negativity; the difficulty lies in realizing a valid profile through $\mathcal{C}$.
Loosely, this is an inverse problem, and solving it by construction requires an architecture that can only represent feasible functions, known only for a few elementary constraint types.
Neural slack variables sidestep this: non-negativity is parameterized exactly on the slack side, while steering the primary network's profile toward the valid target is left to joint training.

\paragraph{Contributions}
\begin{itemize}
    \item We propose \emph{neural slack variables}: a soft constraint method that introduces a non-negative auxiliary network $s^\phi$, jointly trained with the primary network through a matching loss in constraint space.
    \item We identify a drifting failure mode shared by penalty and primal-dual methods, supported by its visual signature in \Cref{fig:drift-monotone} and an ablation that removes its source (\Cref{sec:exp-barrier}), and show that neural slack variables overcome it in the data-driven setting.
    \item We identify regularity transfer from $s^\phi$ to $f^\theta$ as a controllable inductive bias on the constraint profile, enabling constrained learning with spectrally expressive architectures (SIREN, Fourier features).
    \item We demonstrate stronger constraint satisfaction on synthetic monotonicity and convexity benchmarks, and analyze neural slack dynamics on a data-free certification task from neural verification (FOSSIL Barr3).
    \item As the main application, we apply neural slack variables to implied volatility surface modeling and obtain a robustly arbitrage-free generative model, addressing a high-impact industrial challenge.
\end{itemize}

\section{Background and related work}
\label{sec:background}

An extended discussion of related work is provided in \Cref{app:related-work}.

\subsection{Problem setup}
\label{sec:background-setup}

On a bounded domain $\Omega \subset \mathbb{R}^d$, we learn a neural network $f^\theta \colon \Omega \to \mathbb{R}$ from data $\mathcal{D} = \{(x_i, y_i)\}_{i=1}^N$, subject to a functional $m$-dimensional inequality constraint
\begin{equation}\label{eq:bg-constraint}
    \mathcal{C}[f^\theta](x) \geq 0, \quad \text{for all } x \in \Omega,
\end{equation}
where $\mathcal{C}[\cdot]$ is a constraint operator mapping the function to constraint quantities (derivatives, eigenvalues, or other functional properties). 
We write $\mathcal{L}_\text{data}(\theta)$ for the data-fitting loss (e.g., $\mathbb{E}_{(x,y)\sim\mathcal{D}}\|f^\theta(x) - y\|^2$ for the MSE loss).

\subsection{Penalty method}
\label{sec:background-penalty}

The penalty method is the de facto default for enforcing constraints in deep learning.
It augments the data loss with
\begin{equation}\label{eq:bg-hinge}
    %\mathcal{L}_{\text{hinge}, p}(\theta) = \frac{1}{p} \mathbb{E}_{x \sim \mathcal{U}(\Omega)} \!\Big[ \Vert \max\!\big(0,\; \varepsilon - \mathcal{C}[f^\theta](x)\big) \Vert^p_p \Big],
    \mathcal{L}_{\text{hinge}, p}(\theta)
    = \frac{1}{p} \mathbb{E}_{x \sim \mathcal{U}(\Omega)}
    \!\left[
    \left\|\max\!\bigl(0,\; \varepsilon - \mathcal{C}[f^\theta](x)\bigr)\right\|_p^p
    \right],
\end{equation}
where $\mathcal{U}(\Omega)$ is the Uniform distribution over $\Omega$, $p \geq 1$ (typically $p=1$ to encourage sparse violation patterns), the margin $\varepsilon > 0$ encourages strict satisfaction of the constraint, and ``hinge'' signals the use of the positive-part function to compute the constraint residual.
The total loss becomes $\mathcal{L}(\theta) := \mathcal{L}_\text{data}(\theta) + \rho \, \mathcal{L}_{\text{hinge}, p}(\theta)$, where $\rho > 0$ weights the penalty term versus the data loss.
On a grid $\pi = \{x_j\}_{j=1}^n$, this discretizes to the classical exterior penalty method applied to the sampled constraints $\mathcal{C}[f^\theta](x_j) \geq \varepsilon$ \citep[Ch.~17]{nocedalNumericalOptimization2006}.
Here, the level of $\rho$ governs \emph{exactness} (whether the minimizer coincides with the constrained optimum), so $\rho$ is a crucial tuning parameter in practice.
For the linear penalty ($p=1$), exactness may be attained at sufficiently large finite $\rho$, while the quadratic penalty ($p = 2$) is generally exact only in the limit $\rho \to \infty$, motivating schedules that increase $\rho$ over training.
These considerations are, however, largely set aside in deep learning practice; \citet{ramirezPositionAdoptConstraints2025} advocate primal-dual methods instead, citing in particular the burden of penalty weight tuning.

\subsection{Primal-dual methods}
\label{sec:background-lagrangian}

The \emph{primal-dual} or \emph{Lagrangian} formulation of~\eqref{eq:bg-constraint} associates a non-negative multiplier function $\lambda \colon \Omega \to \mathbb{R}^m_{\geq 0}$ with the constraint:
% \begin{equation}\label{eq:bg-lagrangian}
%     \min_\theta \max_{\lambda(\cdot) \geq 0} \; \mathcal{L}_\text{data}(\theta) - \mathbb{E}_{x \sim \mathcal{U}(\Omega)}\!\big[\lambda(x) \, \mathcal{C}[f^\theta](x)\big].
% \end{equation}
\begin{equation}\label{eq:bg-lagrangian}
    \min_\theta \max_{\lambda(\cdot) \geq 0} \; \mathcal{L}_\text{data}(\theta)
    + \mathbb{E}_{x \sim \mathcal{U}(\Omega)}
    \!\left[
        \left\langle \lambda(x),\; \varepsilon - \mathcal{C}[f^\theta](x) \right\rangle
    \right].
\end{equation}
On a grid $\pi = \{x_j\}_{j=1}^n$, this discretizes to one multiplier vector $\lambda_j \in \mathbb{R}_{\geq 0}^m$ per grid point, equivalently one scalar multiplier per grid point and constraint component.
\emph{Gradient descent-ascent} (GDA) on the multiplier vector is the standard primal-dual scheme in constrained deep learning \citep{ramirezPositionAdoptConstraints2025,gallego-posadaCooperLibraryConstrained2025}.
The \emph{augmented Lagrangian} adds a quadratic violation penalty to~\eqref{eq:bg-lagrangian}, convexifying the local saddle and stabilizing GDA~\citep{plattConstrainedDifferentialOptimization1987}; controller-based multiplier updates offer an alternative \citep{stookeResponsiveSafetyReinforcement2020,sohrabiPIControllersUpdating2024}.

When $\Omega$ is high-dimensional or extended by conditioning variables, materializing one scalar per constraint is impractical, motivating direct parametrization of $\lambda$ as a network $\lambda_\phi \colon \Omega \to \mathbb{R}^m_{\geq 0}$, a \emph{neural multiplier}.
\citet{narasimhanApproximateHeavilyconstrainedLearning2020} introduce neural multipliers for heavily-constrained classification, where the multiplier is indexed by a constraint feature vector (directly analogous to the domain coordinate $x$ in our setting).
They study how the capacity of $\lambda_\phi$ distorts the constrained problem: under-parameterized models effectively enforce averaged transformations of the constraint set rather than individual constraints.

\subsection{Architectural constraints}
\label{sec:background-hard}

Finally, it is possible to guarantee feasibility by explicit architecture choices.
Input-Convex Neural Networks~\citep{amosInputConvexNeural2017} enforce convexity via non-negative weights and convex activations; Constrained Monotonic Neural Networks~\citep{runjeConstrainedMonotonicNeural2023,sartorAdvancingConstrainedMonotonic2025} guarantee monotonicity through paired sign-restrictions on the weights and increasing activation functions.
\section{Method}
\label{sec:method}

\begin{figure*}[!t]
    \centering
    \includegraphics[width=\textwidth]{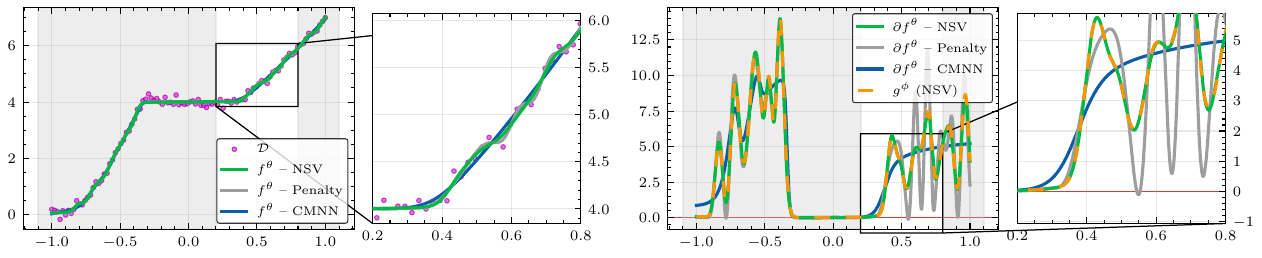}
    \caption{Monotone regression experiment with sinusoidal representation network (SIREN) and fixed constraint grid. Fitted network $f^\theta$ against noisy observations (left) and derivative profile $\partial_x f^\theta$ (right). The penalty method exhibits constraint-grid overfitting (the penalty loss vanishes as violations persist; \Cref{fig:monotone-siren-training-dynamics}), while neural slack variables limit the frequency content of the derivative profile, eliminating violations between collocation points.}
    \label{fig:comparison-monotone-siren}
\end{figure*}

\paragraph{Neural slack variables}

We propose to enforce inequality constraints by augmenting the data loss with the relaxation term $\mathcal{L}_\text{slack}(\theta,\phi)$.
Here, $\phi \in \Phi$ are the parameters of an auxiliary network $s^\phi \colon \Omega \to \mathbb{R}^m$ ($m$ the dimension of the constraint), which is matched to $f^\theta$'s constraint profile:
\begin{equation}
\mathcal{L}_\text{slack}(\theta,\phi) := \frac{1}{2} \mathbb{E}_{x \sim \mathcal{U}(\Omega)}\left[\left\|\mathcal{C}[f^\theta](x) - s^\phi(x)\right\|_2^2\right].
\end{equation}
Crucially, $s^\phi$ is architecturally constrained to take values in $K = [\varepsilon,\infty)^m$ through its output activation function.
We call $s^\phi$ a \emph{neural slack variable}: it carries the feasibility role of a slack variable, while its agreement with $\mathcal{C}[f^\theta]$ is learned through the matching objective rather than imposed exactly.
The total loss becomes $\mathcal{L}(\theta,\phi) := \mathcal{L}_\text{data}(\theta) + \rho\,\mathcal{L}_\text{slack}(\theta,\phi)$, a relaxation of the original constrained problem, minimized jointly in $(\theta,\phi)$.
In practice we replace $\rho\,\mathcal{L}_\text{slack}$ by the relativized loss $\mathcal{L}^{\rho_\text{max}}_\text{slack}$ (\Cref{app:algorithm}), where $\rho_\text{max}$ sets the maximum effective weight near the constraint boundary and no separate multiplier is used.

\paragraph{Gradient persistence via variable splitting}
\label{sec:method-lagrange}

The construction can be understood as variable splitting on the constraint quantity.
Let $c_\theta(x):=\mathcal{C}[f^\theta](x)$ denote the constraint profile of the primary network.
Introduce an auxiliary variable $s$ on $\Omega$ and replace the original constraint $c_\theta \in K$ by the constraint pair $(s = c_\theta) \wedge (s \in K)$.
Variable splitting schemes relax the coupling $s = c_\theta$ (e.g., via penalties or primal-dual methods) and thereby decouple the optimization of the primary model parameters $\theta$ from the formation of a feasible constraint profile $s$.
As the equality residual is reduced, feasibility transfers back to $c_\theta$. 
The simple square penalty variant yields the constrained problem
\begin{align}\label{eq:variable-splitting-objective}
\min_{\theta, s(\cdot)} & \Big\{\mathcal{L}_\text{data}(\theta)
+ \frac{1}{2}\, \mathbb{E}_{x\sim\mathcal{U}(\Omega)} \left[ \left\| c_\theta(x) - s(x) \right\|^2 \right]\Big\}\nonumber \\
 \text{subject to } &
 s(\cdot) \in K.
\end{align}
For fixed $\theta$, the exact $s$-subproblem solved by the pointwise Euclidean projection $\Pi_K(c_\theta(x)) = \max(\varepsilon, c_\theta(x))$.
Substituted into \eqref{eq:variable-splitting-objective} this gives precisely the objective of the quadratic penalty method. %, $\mathcal{L}_\text{data}(\theta)+\mathcal{L}_{\text{hinge},2}(\theta)$.

Instead of optimizing \eqref{eq:variable-splitting-objective} over all feasible auxiliary profiles, we restrict the auxiliary profile to the neural slack class $\mathcal{S} = \{s^\phi \colon \phi \in \Phi \}$ and optimize $(\theta,\phi)$ jointly.
Neural slack variables replace the exact $s$-projection by a learned, finite-capacity, jointly optimized surrogate.
The finite capacity ``approximate projection'' is the key:
The residual $r_{\theta,\phi}(x) = c_\theta(x)-s^\phi(x)$ generally does not collapse to zero throughout the slack region, so that $\mathcal{L}_\text{slack}$ continues to provide an ``anchoring'' gradient even as $c_\theta \geq \varepsilon$, stabilizing feasible constraint profiles and preventing drift.%\footnote{To make the statement more precise, consider the matching gradient, which has finite-sample form $\nabla_\theta \mathcal{L}_\text{slack} = J^\top r / n$, where $J$ is the Jacobian of the sampled constraint profile with respect to $\theta$. The constraint gradient vanishes only when the residual is invisible through the constraint Jacobian.} 

\paragraph{Inductive bias on the constraint profile}

Restricting the auxiliary profile to the finite-capacity class $\mathcal{S}$ adds a term to the inner-minimized slack loss:
\begin{equation}\label{eq:dcr-decomp}
\inf_\phi \mathcal{L}_\text{slack}(\theta,\phi) \;\geq\; \mathcal{L}_{\text{hinge}, 2}(\theta) \;+\; \frac{1}{2}\,\operatorname{dist}_{L^2}^2\!\bigl(\Pi_K (c_\theta),\,\mathcal{S}\bigr).
\end{equation}
The bound is the projection decomposition of $\|c_\theta - s^\phi\|^2$, valid because $\mathcal{S}\subset K$ by construction.
On feasible $\theta$, the quadratic violation penalty term vanishes and the inner-minimized slack loss reduces to $\tfrac{1}{2}\,\operatorname{dist}_{L^2}^2(c_\theta, \mathcal{S})$.
Therefore, the slack loss selects, among feasible $\theta$ of equal data fit, one whose constraint profile lies closest to $\mathcal{S}$.
This is an explicit inductive bias on the shape of $c_\theta$, set by the architecture of $s^\phi$ rather than $f^\theta$.

The strength of the selection depends on the capacity of $\mathcal{S}$.
A small $\mathcal{S}$ imposes a strong shape prior at the cost of underfitting, while excessive $\mathcal{S}$ destabilizes joint optimization.
The useful regime lies between these extremes: $\mathcal{S}$ matched to the constraint profile of $f^\theta$ on the data-relevant subclass.

\section{Experiments}
\label{sec:experiments}

We evaluate neural slack variables against penalty and (augmented) Lagrangian methods; for monotonicity and convexity we additionally report CMNN and ICNN as architectural baselines.
The differential constraint operators we consider amplify high-frequency content, in tension with an MLP-based neural slack class~\citep{rahamanSpectralBiasNeural2019,ramasinghePeriodicityUnifyingFramework2022}.
For the neural slack variable architecture we therefore default to sinusoidal representation networks (SIRENs; \citep{sitzmannImplicitNeuralRepresentations2020}) with input frequency scale $\omega_0 = 5$, biasing the neural slack variable toward smoothness rather than high-bandwidth representation, and ablate this choice in \Cref{sec:exp-monotone} and \Cref{sec:exp-barrier}.
For outputs valued in $K = [\varepsilon, \infty)^m$, we use the componentwise shifted square $\varepsilon + (\cdot)^2$ by default.

We also consider higher bandwidth architectures (moderate SIRENs and Fourier features) for $f^\theta$ in selected experiments; by reducing the low-frequency bias of plain MLPs, these settings arguably better isolate the effect of the constraint-enforcement mechanism.

\paragraph{Training and hyperparameters}
\label{par:training}

With the exception of Experiment~\ref{sec:experiment-vol-autodec}, the datasets are small enough to resolve $\mathcal{L}_\text{data}$ on the full training set (full-batch gradient descent).
All method-specific constraint objectives are evaluated on a separate constraint grid $\pi_\text{train}$, with margin $\varepsilon = 10^{-6}$.
For the penalty method, we use the linear variant ($p=1$).
For penalty and augmented-Lagrangian baselines, we update the scalar penalty coefficient with a progress-based multiplicative rule driven by the maximum constraint violation on $\pi_\text{train}$, capped at a maximum value.
Per-experiment training details (number of epochs, learning rate schedule, network architecture) are reported alongside each benchmark.

\paragraph{Evaluation metrics}
\label{par:eval}

We report the train mean absolute error $\delta_\text{mae}$ and three constraint statistics computed from a fine evaluation grid $\pi_\text{eval}$: the violation rate $\eta_\text{rate}$, the average violation magnitude $\eta_\text{mean}$, and the worst-case magnitude $\eta_\text{max}$ (definitions in \Cref{app:evaluation}).
%We focus on constraint generalization instead of statistical generalization.

\subsection{Testcases: Monotonicity and Convexity}

\subsubsection{\texorpdfstring{Monotonicity ($d=1$)}{Monotonicity (d=1)}}
\label{sec:exp-monotone}

We learn a monotonically increasing function on $\Omega=[-1,1]$ from $100$ noisy observations of a handcrafted target with an adversarial, slightly negatively sloped terrace in the center of the interval (discernible from \Cref{fig:comparison-monotone-siren}; full details in \Cref{app:synthetic-data-monotone}).
The constraint $\mathcal{C}[f](x) = \partial f(x) \geq 0$ is enforced on a $200$-point training grid $\pi_\text{train}$ and verified on a $10{,}000$-point evaluation grid $\pi_\text{eval}$.
We train full-batch for $10{,}000$ epochs with Adam (learning rate $10^{-3}$, cosine-annealed to zero on the final $30\%$).
$f^\theta$ is a width-$16$ four-layer softplus MLP, sized to match the CMNN baseline's minimum depth requirement.

\paragraph{Results}

\Cref{tab:monotone-mlp} aggregates results over $100$ seeds.
Among the soft methods, only neural slack variables achieve zero violations on every seed.
The augmented Lagrangian clears $73/100$ runs against $2/100$ for both the penalty method and the plain Lagrangian; among the latter two, the plain Lagrangian fares strictly worse than the penalty method on every violation metric.
\Cref{fig:drift-monotone} traces the mechanism: the penalty and (augmented) Lagrangian methods continue to drift across the constraint inequality after the data fit converges, while the neural slack variable stabilizes on a strictly non-negative profile.

\begin{table}[h]
    \centering
    \caption{Monotone (MLP $f^\theta$): mean $\pm$ std over $N = 100$ seeds; $n_\text{sat}/N$ counts seeds with $\eta_\text{rate}=0$. The neural slack variable induces feasibility on every seed. Apple M3 Pro (CPU).}
    \label{tab:monotone-mlp}
    \resizebox{\columnwidth}{!}{%
    \begin{tabular}{lllllll}
\toprule
Method & $\delta_\text{mae}$ & $\eta_\text{rate}$ & $\eta_\text{mean}$ & $\eta_\text{max}$ & $n_\text{sat}/N$ & Time (s) \\
\midrule
Baseline & 7.8e-02 ± 6.0e-03 & 2.4e-01 ± 2.6e-02 & 1.4e-01 ± 2.6e-02 & 1.2e+00 ± 3.8e-01 & 0/100 & 22.8 \\
Penalty & 9.2e-02 ± 7.0e-03 & 6.0e-03 ± 3.0e-03 & 1.1e-06 ± 1.2e-06 & 3.2e-04 ± 3.1e-04 & 2/100 & 32.4 \\
Lagrangian & 9.1e-02 ± 7.3e-03 & 2.0e-02 ± 9.4e-03 & 1.9e-05 ± 1.7e-05 & 1.7e-03 ± 1.4e-03 & 2/100 & 29.4 \\
Aug. Lag. & 9.3e-02 ± 7.1e-03 & 4.8e-03 ± 9.2e-03 & 2.8e-06 ± 6.1e-06 & 2.3e-04 ± 4.5e-04 & 73/100 & 32.9 \\
N. Slack Var. & 9.6e-02 ± 6.8e-03 & 0 & 0 & 0 & 100/100 & 44.7 \\
CMNN & 8.5e-02 ± 6.7e-03 & 0 & 0 & 0 & 100/100 & 27.6 \\
\bottomrule
\end{tabular}
}
\end{table}
    
Among the feasible methods (\Cref{tab:monotone-mlp} above), CMNN reaches the best $\delta_\text{mae}$ with an MLP $f^\theta$ while, in this configuration, the neural slack variable's constraint enforcement costs the most accuracy.
In the following two paragraphs we investigate the role of the neural slack architecture as well as the choice of the primary network on these metrics.

\paragraph{Ablation: Neural slack variable architecture and capacity}
\label{sec:exp-monotone-capacity}

\begin{figure}[!t]
\centering
\begin{subfigure}[b]{0.59\columnwidth}
    \centering
    \includegraphics[width=\columnwidth]{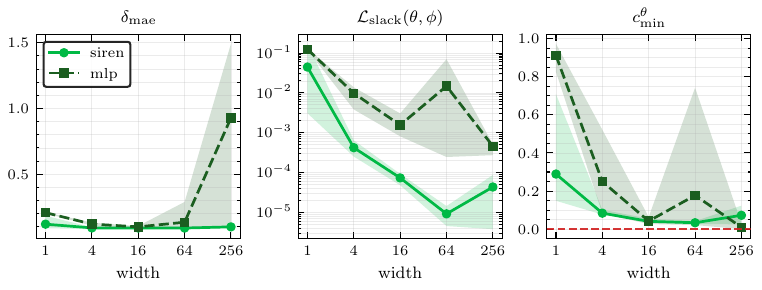}
    \subcaption{Monotonicity: Neural slack variable, hidden width sweep.}
    \label{fig:monotone-aux-capacity-sweep}
\end{subfigure}\hfill
\begin{subfigure}[b]{0.39\columnwidth}
    \centering
    \includegraphics[width=\columnwidth]{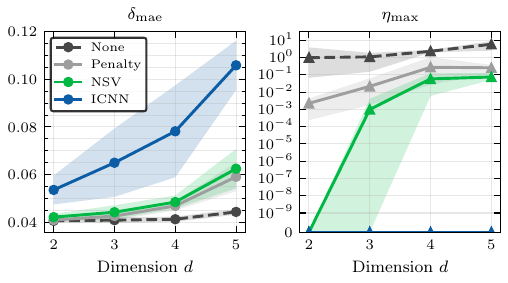}
    \subcaption{Convexity: Increasing dimension}
    \label{fig:convex-nd-scaling}
\end{subfigure}
\caption{\textbf{(a)} Neural slack variable width sweep on the monotone MLP $f^\theta$, comparing a SIREN slack against a softplus-MLP slack. Low neural slack capacity hurts $\delta_\text{mae}$ for the MLP slack disproportionally; excessive capacity destabilizes the MLP slack at $W=256$. The SIREN slack stays flat across the sweep; both choices keep $c^\theta_{\min}$ positive at every width. \textbf{(b)} Convex-ND (MLP $f^\theta$), $5$ seeds: data MAE (left) and worst-case violation $\eta_\text{max}$ (right) versus input dimension $d$, comparing the neural slack variable against the penalty method, an unconstrained reference, and the architecturally constrained ICNN. Bands are $\pm 1$ std for $\delta_\text{mae}$ and seed min/max for $\eta_\text{max}$. Feasibility degrades with $d$ as the constraint training grid thins for the penalty method and neural slack variable; the ICNN remains feasible by construction but trails on $\delta_\text{mae}$ at every $d$.}
\label{fig:capacity-and-scaling}
\end{figure}

The neural slack variable used for \Cref{tab:monotone-mlp} is a SIREN with $\omega_0 = 5$ and three hidden layers of width 16.
We ablate this choice and by contrasting with $\mathrm{softplus}$-MLPs and sweeping both architectures' widths.
\Cref{fig:monotone-aux-capacity-sweep} reports $\delta_\text{mae}$, $\mathcal{L}_\text{slack}(\theta, \phi)$ and $c^\theta_{\min}$ across $5$ seeds; constraint satisfaction is exact at every width for both architectures.

The SIREN slack attains a lower $\delta_\text{mae}$ than the MLP slack across the full sweep, with the gap closing at $W = 16$.
Both architectures exhibit a U-shape in $\delta_\text{mae}$: capacity is needed at the lower end, where a narrow neural slack variable overly restricts the expressivity of the primary network, and excess capacity destabilises the optimization.
The failure modes at $W = 256$ differ in quality (\Cref{fig:monotone-aux-derivatives,fig:monotone-aux-derivatives-siren}).
The $\mathrm{softplus}$-MLP slack collapses to zero, which forces $f^\theta$ itself toward a constant predictor and raises $\delta_\text{mae}$ above one.
The SIREN slack instead settles into a pathological microscopic-oscillation state that still traces a data-compatible profile on a macroscopic level, so $\delta_\text{mae}$ rises modestly without catastrophic collapse.

\paragraph{\texorpdfstring{Using a SIREN for $f^\theta$}{Using a SIREN for f-theta}}

Using a width-$16$ three-layer SIREN for $f^\theta$ widens the spectral envelope available to the regression itself.
\Cref{fig:comparison-monotone-siren} contrasts $f^\theta$ and $\partial_x f^\theta$ in this setting.
The fixed-grid run exposes constraint-grid overfitting for the penalty method: its penalty loss term vanishes while violations persist (\Cref{fig:monotone-siren-training-dynamics}), whereas the neural slack variable selects a lower-frequency derivative profile that remains positive between collocation points.
The corresponding table in \Cref{app:monotone-siren-results} reports $\delta_\text{mae} = 0.071$ for the neural slack variable against $0.081$ for CMNN, with both feasible on every seed, whereas the penalty method leaves substantial residual violations and the Lagrangian methods suffer failures, with primal-dual updates struggling to handle the SIREN's constraint profile.
Using neural slack variables moves constraint enforcement out of the primary architecture, freeing enough capacity for $f^\theta$ to overtake CMNN on $\delta_\text{mae}$ while preserving feasibility in every run.

\subsubsection{\texorpdfstring{Convexity ($d \in \{2, 3, 4, 5\}$)}{Convexity (d in {2, 3, 4, 5})}}

\label{sec:exp-convex-nd}

We learn convex functions on $\Omega = [-1, 1]^d$ for $d \in \{2, 3, 4, 5\}$ from a smoothed log-sum-exp target with $K = 10$ random affine pieces (full details in \Cref{app:synthetic-data-convex-nd}).
The constraint $\mathcal{C}[f](x) = \operatorname{eig} \nabla^2 f(x) \geq 0$ requires every Hessian eigenvalue to be non-negative.
We train full-batch with Adam on $N = 10^3$ noisy observations for $10{,}000$ epochs (learning rate $10^{-3}$, cosine-annealed to zero on the final $30\%$).
At each optimizer step, the constraint objective is evaluated on a newly drawn Sobol grid $\pi_\text{train}$ with $|\pi_\text{train}| = 10^4$.
Evaluation uses an independent Sobol grid $\pi_\text{eval}$ with $|\pi_\text{eval}| = 10^7$.
The primary model $f^\theta$ is a width-$128$ three-layer $\mathrm{softplus}$-MLP, matching the activation class required by the ICNN baseline; the neural slack variable $s^\phi$ is a width-$128$ three-layer SIREN with input frequency $\omega_0 = 5$.
We run $5$ seeds per dimension, alongside an unconstrained reference.
Per-dimension tables and $d=2$ surface plots are in \Cref{app:convex-nd}.

\paragraph{Results}

\Cref{fig:convex-nd-scaling} shows the scaling across input dimension; bands are $\pm 1$ std for $\delta_\text{mae}$ and seed min/max for $\eta_\text{max}$, so an $\eta_\text{max}$ band whose lower edge is at zero contains at least one feasible seed.
At $d=2$ neural slack variables ensure feasibility on every seed, while the penalty method fails on every seed ($\eta_\text{max} \approx 2 \times 10^{-3}$).
At $d=3$ neural slack variables start to leak violations for some seeds, while the penalty method grows by an order of magnitude ($\eta_\text{max} \approx 2 \times 10^{-2}$); at $d=4,5$ neither method clears any seed, and the penalty's worst-case violation is three to five times the slack variable's.
$\delta_\text{mae}$ remains comparable between both methods at every $d$.
We read this as the curse of dimensionality acting on learned constraint enforcement: at fixed grid budget the per-volume sample density thins exponentially, and methods whose feasibility relies on dense pointwise supervision degrade accordingly.
ICNNs are convex by construction but underfit at $d = 2, 3$, consistent with the slower training dynamics documented for input-convex architectures~\citep{sivaprasadCuriousCaseConvex2021,hoedtPrincipledWeightInitialisation2023}.

\subsection{Neural barrier certificates}
\label{sec:exp-barrier}

We turn to a multi-component safety certificate drawn from the FOSSIL Barr3 benchmark~\citep{edwardsFossil20Formal2024}.
On the autonomous polynomial system $\dot{x} = \Phi(x)$ with $\Phi(x) = (x_2,\; -x_1 - x_2 + \tfrac{1}{3}\,x_1^{3})$ over $\Omega = [-3, 2.5] \times [-2, 1]$, we train $f^\theta \colon \Omega \to \mathbb{R}$ whose certified-safe set $\{x : f^\theta(x) \geq 0\}$ separates an initial set $\Omega_\mathrm{i}$ from an unsafe set $\Omega_\mathrm{u}$, both unions of axis-aligned rectangles and disks (full geometry in \Cref{app:barrier-setup}).
Folding the boundary sign anchors and the Nagumo / control-barrier inequality~\citep{amesControlBarrierFunctions2019,dawsonSafeControlLearned2023} into a single vector-valued constraint:
\begin{equation}\label{eq:barrier-c}
    \resizebox{0.9\linewidth}{!}{$
    \mathcal{C}[f^\theta](x)
    =
    \bigl(
    \nabla f^\theta(x)\cdot\Phi(x) + \gamma\, f^\theta(x),\;
    \mathbf{1}_{\Omega_\mathrm{i}}(x)\bigl(f^\theta(x)-\mu_\mathrm{i}\bigr),\;
    \mathbf{1}_{\Omega_\mathrm{u}}(x)\bigl(-f^\theta(x)-\mu_\mathrm{u}\bigr)
    \bigr)
    \;\ge\; 0,
    $}
\end{equation}
with $\gamma = 1$ and strict-positivity margins $\mu_\mathrm{i} = \mu_\mathrm{u} = 10^{-3}$ on the boundary anchors so that the certificate's safe set has non-empty interior on $\Omega_\mathrm{i}$ and a strict separation from $\Omega_\mathrm{u}$.
We give details about training and evaluation grids in \Cref{app:barrier-setup}.

This setting removes the data-fitting term ($\mathcal{L}_\text{data} \equiv 0$), leaving the constraint-enforcement term as the only signal for $\theta$.
It therefore removes the source of constraint drift identified in \Cref{fig:drift-monotone}.
Indeed, the literature-default two-layer width-$10$ squared-activation MLP~\citep{edwardsFossil20Formal2024,demouraZ3EfficientSMT2008} reaches feasibility within about $2{,}000$ gradient steps under the penalty method; at this point, with no data term and zero penalty loss, its parameters freeze.
This setting is therefore the controlled counterpart to \Cref{fig:drift-monotone}: the penalty method suffers from fragile constraint satisfaction under a competing data fit, but not in its absence.
This setting is therefore the controlled counterpart to \Cref{fig:drift-monotone}: the penalty method exhibits fragile constraint satisfaction under a competing data fit, but not when the data term is absent.

\begin{figure}[!t]
\centering
\includegraphics[width=\columnwidth]{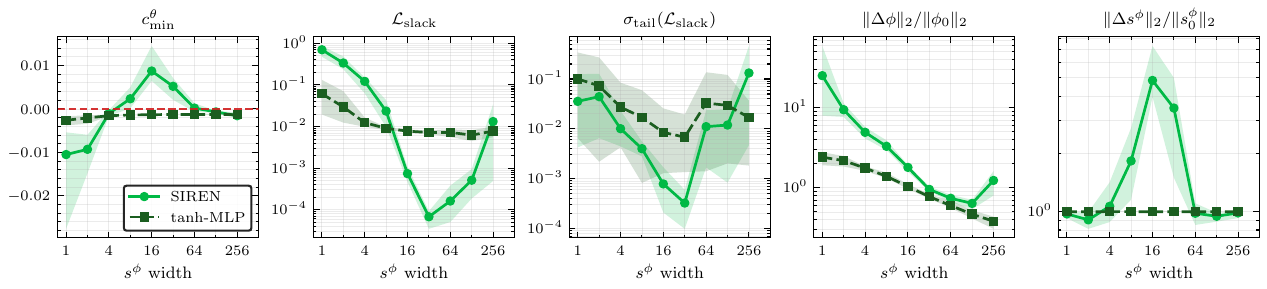}
\caption{Neural slack variable width sweep on FOSSIL Barr3, $5$ seeds per cell (lines = seed mean, bands = seed min/max). Linear panel: $c^\theta_{\min}$ (dashed line marks feasibility; the band lower edge is the worst-seed feasibility). Log panels: slack loss $\mathcal{L}_\text{slack}$, its tail oscillation $\sigma_\tau$, the relative parameter displacement $\|\Delta\phi\|_2 / \|\phi_0\|_2$, and the relative output displacement $\|\Delta s^\phi\|_2 / \|s^\phi_0\|_2$ on a fixed grid. SIREN is feasible only at intermediate widths $h \in \{8, 16, 32\}$; the $\tanh$-MLP slack is infeasible at every width tried.}
\label{fig:barrier-aux-capacity}
\end{figure}

\paragraph{Neural slack variable dynamics}
\label{sec:exp-barrier-capacity}

For neural slack variables, the absence of a data term makes Barr3 a diagnostic of the method's own training dynamics.
\Cref{fig:barrier-aux-capacity} sweeps the hidden width of a three-layer neural slack variable $s^\phi$ over $\{1, 2, \ldots, 256\}$ for SIREN ($\omega_0 = 5$) and $\tanh$-MLP architectures, trained full-batch for $5{,}000$ epochs (details in \Cref{app:barrier-knobs}).
Feasibility requires $s^\phi$ to represent an inflated, sink-shaped constraint profile, even though the certificate $f^\theta$ itself remains small in magnitude.
\Cref{fig:barrier-aux-capacity} identifies the collapse regime: since the neural slack variable is initialized at constant unit output, $s^\phi_0 \equiv 1$, collapse toward zero gives $\|\Delta s^\phi\|_2 / \|s^\phi_0\|_2 \approx 1$ (rightmost panel).
The $\tanh$-MLP slack variable remains in this regime at every width, failing to induce feasibility.
The SIREN slack variable leaves this regime only at intermediate widths $h \in \{8, 16, 32\}$, with peak feasibility at $h = 16$.
The tail oscillation $\sigma_\tau$ (standard deviation of $\mathcal{L}_\text{slack}$ over the last $1{,}500$ epochs) is lowest in the same intermediate-width regime and rises at both extremes.
The high-width end aligns with the capacity failure seen in the monotone ablation (\Cref{fig:monotone-aux-derivatives-siren}), where $s^\phi$ appears to move through SIREN feature combinations.
The low-width end suggests that joint training of $f^\theta$ and $s^\phi$ fails to settle into a stable configuration, with no data-fitting gradient to anchor it.

\subsection{No-arbitrage volatility surfaces}
\label{sec:experiment-vol-autodec}

Constructing arbitrage-free implied volatility surfaces is a longstanding challenge in equity-options trading: even standard low-dimensional parametrizations \citep{gatheralArbitragefreeSVIVolatility2014,lucicNormalizingVolatilityTransforms2021} require careful constraint handling.
Deep-learning surrogates predominantly enforce no-arbitrage softly through penalty methods \citep{ackererDeepSmoothingImplied2020,chataignerDeepLocalVolatility2020,zhengIncorporatingPriorFinancial2021,bergeronVariationalAutoencodersHandsoff2022,yangHyperIVRealtimeImplied2025,wiedemannOperatorDeepSmoothing2025}; satisfaction rates are either reported as imperfect or not reported at all, implying residual arbitrage at deployment.
Architecturally constrained models are arbitrage-free by construction, but their mismatch with market data is well documented and has limited their adoption in practice \citep{chataignerDeepLocalVolatility2020}.
We adopt this as the paper's main application: $251$ daily NDX (Nasdaq-100) option snapshots from OptionMetrics, fit jointly through a variational autodecoder \citep{zadehVariationalAutoDecoderMethod2021} that maps a per-surface latent code $z_i \in \mathbb{R}^{32}$ and grid coordinates $(\tau, k)$ (time-to-expiry and log-moneyness) to implied volatility through a FiLM-ed SIREN $f^\theta(\tau, k; z_i)$ (\citet{perezFiLMVisualReasoning2018}; architecture details in \Cref{app:vol-autodec-model}).

Writing $v(\tau, k) := f^\theta(\tau, k; z)\sqrt{\tau}$ for \emph{total volatility} and $w := v^2$ for \emph{total variance}, the no-arbitrage constraint is two-dimensional:
\begin{equation}\label{eq:vol-c}
    %\resizebox{0.9\linewidth}{!}{$
    \mathcal{C}[f^\theta]
    =
    \bigl(
    \mathcal{C}_\text{cal}[f^\theta],\;
    \mathcal{C}_\text{str}[f^\theta]
    \bigr)
    =
    \bigl(
    \partial_\tau w,\;
    (1 + d_1\,\partial_k v)(1 + d_2\,\partial_k v) + v\,\partial_k^2 v
    \bigr)
    \;\ge\; 0,
    %$}
\end{equation}
where $d_{1,2}(\tau, k, v) := -k/v \pm v/2$ (details in \Cref{app:vol-autodec-options}).
The \emph{local volatility} $a(\tau, k)$ derived from $\mathcal{C}[f^\theta]$ via Dupire's equation ($a^2 = \mathcal{C}_\text{cal}/\mathcal{C}_\text{str}$) is a standard diffusion model for pricing exotic derivatives consistently with the observed option prices; residual violations of either constraint leave $a$ undefined.
Deployment-grade surfaces therefore require strict enforcement of both constraints.

The setting introduces a qualitatively new challenge: a high-dimensional, structured constraint domain.
The shape constraint targets the surface coordinates $(\tau, k) \in [\tau_\text{min}, \tau_\text{max}] \times [k_\text{min}, k_\text{max}]$ but is extended pointwise by conditioning on the latent, so the effective constraint domain is $\Omega \subseteq \mathbb{R}^{34}$ with $x = (\tau, k, z)$.
Mini-batches of $8$ surfaces per step further make the constraint grid seen by each update very sparse, so the constraint enforcement method must interpolate cleanly through both the spatial and the latent dimensions under that noise.
Finally, feasibility must hold beyond the trained codes $\{z_i\}_i$ (\textsc{train}) to make the generative setting useful; we additionally evaluate it on linear interpolations of pairwise sampled trained codes (\textsc{interpolation}) and at samples from the latent prior (\textsc{prior}), the deployment regime for unconditional sampling.

\paragraph{Results}
\Cref{tab:vol-autodec} compares the penalty method, the neural augmented Lagrangian, and neural slack variables.
Neural slack variables are the only method to fully eliminate violations across all three regimes (\textsc{train}, \textsc{interpolation}, and the deployment-regime \textsc{prior}).
Moreover, it achieves the best data fit; the augmented Lagrangian attains the next-best fit but the largest residual violations across all regimes.
Neural slack variables thereby establish a viable approach to arbitrage-free learning of volatility surfaces, in our view an open practical challenge of high industrial impact.

\begin{table}[!t]
\centering
\caption{IV-surface autodecoder: method comparison. Lower is better. Only the neural slack variable is feasible on every regime, at the lowest fit error. Measured on Apple M3 Pro (MPS).}
\label{tab:vol-autodec}
\resizebox{\columnwidth}{!}{%
\begin{tabular}{lcccccccc}
\toprule
&  & \multicolumn{2}{c}{\textsc{train}} & \multicolumn{2}{c}{\textsc{pairwise}} & \multicolumn{2}{c}{\textsc{prior}} &  \\
\cmidrule(lr){3-4} \cmidrule(lr){5-6} \cmidrule(lr){7-8}
Method & $\delta_\text{mae}$ & $\eta_\text{rate}$ & $\eta_\text{max}$ & $\eta_\text{rate}$ & $\eta_\text{max}$ & $\eta_\text{rate}$ & $\eta_\text{max}$ & Time (h) \\

\midrule
Penalty & 1.0e-02 & 5.6e-04 & 2.7e-03 & 3.3e-04 & 1.6e-03 & 4.9e-04 & 2.8e-03 & 1.1 \\
Neural Aug. Lag. & 7.8e-03 & 1.2e-03 & 4.9e-03 & 1.1e-03 & 4.4e-03 & 1.5e-03 & 7.7e-03 & 1.2 \\
Neural Slack Var. & 6.9e-03 & 0 & 0 & 0 & 0 & 0 & 0 & 1.3 \\
\bottomrule
\end{tabular}

}
\end{table}

\section{Conclusion}
\label{sec:discussion}

Neural slack variables enforce shape constraints by pairing the primary network with an auxiliary network that is feasible by construction and jointly trained as a target for the constraint profile of $f^\theta$.
The construction overcomes the drifting failure mode of penalty and primal-dual methods (\Cref{fig:drift-monotone}); removing the data loss in the barrier experiment isolates the cause (\Cref{sec:exp-barrier}).
On dense-grid synthetic benchmarks the method achieves zero measured violations on every seed; with a spectrally expressive primary network (SIREN, Fourier features) it maintains constraint enforcement while achieving a lower data error than the architectural baselines, a regime not reached by the penalty or primal-dual baselines in our experiments.
On the high-dimensional volatility surface application (\Cref{sec:experiment-vol-autodec}), a high-profile open challenge in deep learning for quantitative finance, it is the only method to satisfy the no-arbitrage constraints on every evaluation regime while attaining the lowest data fit error of the three methods.

The neural slack architecture is an explicit inductive bias on the constraint profile, with a U-shaped capacity tradeoff. Because differential constraint operators amplify high frequencies, the neural slack variable typically needs more spectral expressiveness than $f^\theta$.

\paragraph{Limitations}

Learned constraint enforcement on a sampled grid scales poorly with input dimension; neural slack variables degrade more slowly than the penalty method but do not avoid this scaling issue (\Cref{sec:exp-convex-nd}).
The neural slack variable roughly doubles parameter count and adds an architectural design axis the penalty method does not carry, though dense grid constraint evaluation dominates the per-step cost in either case.
Off-grid feasibility is not formally certified, so architectural constraints remain the appropriate default where available and applicable.

% Acknowledgements are hidden during review by the ack environment
\begin{ack}
AJ gratefully acknowledges financial support from the EPSRC grants EP/W032643/1 and EP/T032146/1.
RW was supported by the Department of Mathematics, Imperial College, through the Roth scholarship scheme.
Computational resources and support were provided by the Imperial College Research Computing Service \citep{harveyImperialCollegeResearch2017}.
For the purpose of open access, the authors have applied a Creative Commons Attribution (CC BY) licence to any Author Accepted Manuscript version arising from this work.
\end{ack}

\bibliography{library2}
\bibliographystyle{plainnat}

\newpage
\appendix

\let\dcrappendixsection\section
\renewcommand{\section}{\clearpage\dcrappendixsection}

\section{Neural slack variables: training and implementation details}
\label{app:algorithm}

\begin{algorithm}[h]
    \caption{Neural slack variable training}
    \label{alg:nsv}
    \begin{algorithmic}[1]
        \STATE \textit{Input:} dataset $\mathcal{D}$; primary network $f^\theta \colon \Omega \to \mathbb{R}$; neural slack variable $s^\phi \colon \Omega \to K$; constraint operator $\mathcal{C}$; data batch size $B_\text{d}$, grid batch size $B_\text{g}$; maximum effective weight $\rho_\text{max}$; iterations $T$.
        \STATE \textit{Initialize:} $\theta^{(0)}, \phi^{(0)}$; set $\delta \gets 1/\sqrt{\rho_\text{max}}$.
        \FOR{$t = 1,\dots,T$}
            \STATE Sample $\{(x_n, y_n)\}_{n=1}^{B_\text{d}} \subset \mathcal{D}$ and $\{\xi_n\}_{n=1}^{B_\text{g}} \sim \mathcal{U}(\Omega)$.
            \STATE Evaluate $c_n \gets \mathcal{C}[f^\theta](\xi_n)$ and $s_n \gets s^\phi(\xi_n)$.
            \STATE $d_n \gets \max\bigl(\delta,\ \mathrm{sg}[s_n]\bigr)$.
            \STATE $\mathcal{L}_\text{data} \gets \tfrac{1}{B_\text{d}} \sum_{n=1}^{B_\text{d}} \ell\bigl(f^\theta(x_n), y_n\bigr)$.
            \STATE $\mathcal{L}_\text{slack} \gets \tfrac{1}{2 B_\text{g}} \sum_{n=1}^{B_\text{g}} \bigl\|\operatorname{asinh}(c_n/d_n) - \operatorname{asinh}(s_n/d_n)\bigr\|^{2}$.
            \STATE Optimizer step on $\mathcal{L}_\text{data} + \mathcal{L}_\text{slack}$ jointly in $(\theta, \phi)$.
        \ENDFOR
    \end{algorithmic}
\end{algorithm}

\paragraph{Scale-balanced slack matching}

A plain squared matching loss between $c_\theta = \mathcal{C}[f^\theta]$ and $s^\phi$ is dominated by satisfied margins: inequality constraints typically have wide dynamic range, and large values of $c_\theta$ contribute uninformative gradient.
We instead match $c_\theta$ and $s^\phi$ on a $s^\phi$-relative scale.
The choice has a self-consistent justification.
Let $\hat{c}$ denote the constraint profile of the feasible solution selected by the data and the inductive bias of $\mathcal{S}$.
Late in training one expects $s^\phi \approx \hat{c}$, so $s^\phi$ is itself an estimator of the natural denominator $\hat{c}$.
Where $\hat{c}$ is small, the relative metric weights residuals heavily, exactly the regime where drift can occur; where $\hat{c}$ is large, deviations are penalized in proportion, so satisfied margins do not dominate.
The architectural lower bound $s^\phi \ge \varepsilon$ also keeps the denominator strictly positive.

\paragraph{Adaptive interpretation}

The relative scaling caps the effective penalty weight at a chosen $\rho_\text{max}$.
With $\delta := 1/\sqrt{\rho_\text{max}}$ and
\begin{equation}\label{eq:denom-floor}
    d_\phi(x) := \max\bigl(\delta,\ \mathrm{sg}[s^\phi(x)]\bigr),
\end{equation}
where $\mathrm{sg}[\,\cdot\,]$ is a stop-gradient on $s^\phi$, the relativized matching loss
\begin{equation}\label{eq:rel-slack-loss}
\mathcal{L}_\text{slack}^{\rho_\text{max}}(\theta,\phi)
\;:=\; \frac{1}{2}\,\mathbb{E}_{x \sim \mathcal{U}(\Omega)}\!\left[\biggl\|\frac{c_\theta(x) - s^\phi(x)}{d_\phi(x)}\biggr\|^{2}\right]
\end{equation}
satisfies $\mathcal{L}_\text{slack}^{\rho_\text{max}} = \rho_\text{max}\,\mathcal{L}_\text{slack}$ in the binding regime $s^\phi \le \delta$, and is scale-invariant for $s^\phi > \delta$.
There is therefore no separate multiplier on $\mathcal{L}_\text{slack}$ in the practical algorithm; $\rho_\text{max}$ sets the maximum effective quadratic weight near the constraint boundary.

\paragraph{Asinh stabilization}

For additional training stability, we apply an $\operatorname{asinh}$ transform to the rescaled arguments before differencing, replacing the squared residual in \eqref{eq:rel-slack-loss} by
\begin{equation}\label{eq:rel-slack-loss-asinh}
\biggl\|\operatorname{asinh}\!\biggl(\frac{c_\theta(x)}{d_\phi(x)}\biggr) - \operatorname{asinh}\!\biggl(\frac{s^\phi(x)}{d_\phi(x)}\biggr)\biggr\|^{2}.
\end{equation}
$\operatorname{asinh}$ is locally the identity at the origin, so the binding-regime equivalence is preserved to leading order.
We found the transform beneficial in our experiments.

\Cref{alg:nsv} summarizes the joint training procedure for the primary network $f^\theta$ and the neural slack variable $s^\phi$.

\section{Related work}
\label{app:related-work}

\paragraph{Theory}
The PAC-learning framework gives distributional guarantees for constrained learning through Lagrangian duality~\cite{chamonProbablyApproximatelyCorrect2020,chamonConstrainedLearningNonconvex2023}.
In this setting, \cite{hounieResilientConstrainedLearning2023} study adaptive constraint relaxation, with slack adjusted during training according to a user-specified relaxation cost.
Neural slack variables also relax constraints, but through a learned slack network rather than per-sample slack updates.

\paragraph{Constrained optimization learning}
A related line of work uses neural networks to amortize finite-dimensional constrained optimization, where the network learns a surrogate solver from problem instances to optimal decisions~\cite{kotaryEndtoendConstrainedOptimization2021}.
Our setting instead trains one function under functional inequality constraints, often involving derivatives, over a continuous domain.
\citet{parkSelfsupervisedPrimaldualLearning2023} also learn primal and dual networks for instance-amortized constrained optimization.
The closest connection to our method is their target-based dual training, where the dual network is trained toward a projected update rather than only through a violation penalty.
Neural slack variables make a related move on the primal side: they learn an auxiliary target and match the constraint operator to it, giving gradient signal beyond the currently violated set.

\paragraph{Projection-based hard constraint methods}

Several recent works enforce hard constraints through differentiable projection or solver layers~\cite{minHardNetHardConstrainedNeural2025,iftakherPhysicsinformedNeuralNetworks2026,golderDAEHardNetPhysicsConstrained2025}.
These methods resemble neural slack variables because both form explicit feasible targets, but they address different settings.
Projection-based methods typically use fixed, non-learnable operators and often require an optimization or solve inside the forward pass.
Neural slack variables instead target functional inequality constraints with a learnable slack network, where the joint learning of the matching objective replaces the explicit projection.

Related neuro-symbolic work also uses symbolic constraint descriptions to enforce linear inequalities in generative models for tabular data~\cite{stoianHowRealisticYour2024}.
Such approaches suggest a route toward more general hard-constraint mechanisms, which remain limited for many applied constrained learning problems.

\paragraph{Sobolev training}
When the constraint operator $\mathcal{C}[\cdot]$ involves derivatives (monotonicity via first derivatives, convexity via Hessian spectra), neural slack variables are related to Sobolev training \cite{czarneckiSobolevTrainingNeural2017, choSobolevTrainingOperator2025}.
Both introduce derivative-based matching terms, but neural slack variables learn the target profile $s^\phi$ jointly with $f^\theta$ rather than taking it from derivative labels.

\clearpage

\section{Experimental evaluation metrics}
\label{app:evaluation}

We report approximation quality via mean absolute error
\begin{equation*}
\delta_\text{mae}
:=
\mathbb{E}_{(x,y)\sim\mathcal{D}}\!\left[\left|f^\theta(x)-y\right|\right],
\end{equation*}
computed on the full training dataset $\mathcal{D}$.

Constraint satisfaction is measured on a fine evaluation grid $\pi_\text{eval}$ covering~$\Omega$.
For a constraint operator $\mathcal{C}[f^\theta] \colon \Omega \to \mathbb{R}^m$ with components $\mathcal{C}_j[f^\theta]$, $j = 1, \dots, m$, we report
\begin{align*}
\eta_\text{rate}
&:=
\mathbb{E}_{x \sim \pi_\text{eval}}\!\left[\mathbf{1}_{\left\{\mathcal{C}[f^\theta](x) \not\geq 0\right\}}\right],
\\
\eta_\text{mean}
&:=
\mathbb{E}_{x \sim \pi_\text{eval}}\!\left[\frac{1}{m}\sum_{j=1}^{m} \max\!\left\{-\mathcal{C}_j[f^\theta](x),\, 0\right\}\right],
\\
\eta_\text{max}
&:=
\max_{x \in \pi_\text{eval}}\,\max_{j = 1, \dots, m}\, \max\!\left\{-\mathcal{C}_j[f^\theta](x),\, 0\right\},
\end{align*}
where $\mathbf{1}_{\{\cdot\}}$ is the indicator function.
$\eta_\text{rate}$ is the fraction of evaluation points where (any component of) the inequality constraint is violated; $\eta_\text{mean}$ averages the elementwise shortfall over both the evaluation grid and the constraint components, so non-violating components contribute zero; $\eta_\text{max}$ is the worst-case shortfall.
For seed-aggregated tables we additionally report $n_\text{sat}/N$, the count of seeds with $\eta_\text{rate} = 0$ exactly.

\section{\texorpdfstring{Additional details: Monotonicity ($d=1$)}{Additional details: Monotonicity (d=1)}}

\subsection{Dataset and target function}
\label{app:synthetic-data-monotone}

The target is the squared piecewise linear profile
\begin{equation*}
    f(x) = b(x)^2, \qquad
    b(x) = \begin{cases}
        (2 - s/3) + 3\,(x + \tfrac{1}{3}), & -1 \leq x < -\tfrac{1}{3}, \\[2pt]
        2 + s\,x, & -\tfrac{1}{3} \leq x < \tfrac{1}{3}, \\[2pt]
        (2 + s/3) + (x - \tfrac{1}{3}), & \tfrac{1}{3} \leq x \leq 1,
    \end{cases}
\end{equation*}
on $\Omega = [-1, 1]$, where the inner-slope parameter $s$ controls the central segment.
The two outer segments retain slopes~$3$ (left) and~$1$ (right); their vertical anchors are chosen so $b$ is continuous at $x = \pm \tfrac{1}{3}$ for any $s$.
With $s = 0$ the central segment is a flat plateau at $f \equiv 4$.
We use $s = -0.1$, tilting the plateau into a slight downward terrace; this violates $\partial_x f \ge 0$ in the ground truth, so the constraint actively opposes the data fit.

The training dataset consists of $N = 100$ noisy observations on a uniform grid over $\Omega$, with additive zero-mean Gaussian noise (standard deviation $0.1$).
The constraint is enforced on a separate uniform training grid $\pi_\text{train}$ of $200$ points covering $[-1.05, 1.05]$ (a small $0.05$ pad past the data domain) and verified on an evaluation grid $\pi_\text{eval}$ of $10{,}000$ uniformly spaced points over $\Omega$.
Initialization seeds and noise seeds are paired across runs; exact reproduction is possible from the supplemental code.

\subsection{Architectural baseline (CMNN)}
\label{app:monotone-cmnn}

As an architectural reference we use (advanced) Constrained Monotonic Neural Networks (CMNNs)~\cite{sartorAdvancingConstrainedMonotonic2025}.
Following the reference implementation,\footnote{\url{https://github.com/AMCO-UniPD/monotonic}} the first hidden-layer activation is the identity and subsequent layers use a monotone nonlinearity; we use $\operatorname{Softplus}$.
We match the primary network's capacity: four hidden layers of width $16$, with bias terms enabled.

\subsection{\texorpdfstring{SIREN $f^\theta$ results}{SIREN f-theta results}}
\label{app:monotone-siren-results}

Counterpart to \Cref{tab:monotone-mlp} with $f^\theta$ and $s^\phi$ both three-layer SIRENs of width $16$ and $\omega_0 = 15$; other settings match \Cref{sec:exp-monotone}.

\begin{table}[h]
\centering
\caption{Monotone (SIREN $f^\theta$): mean $\pm$ std over $10$ seeds. The neural slack variable achieves a lower $\delta_\text{mae}$ than CMNN at full feasibility (10/10 seeds): using a SIREN for $f^\theta$ extracts more variance from the data than CMNN can, with the neural slack variable still enforcing feasibility. Counterpart to \Cref{tab:monotone-mlp}; the penalty method still leaves small residual violations, while the Lagrangian methods fail, with primal-dual updates struggling to handle the SIREN's constraint profile (the augmented Lagrangian method was feasible on 73/100 seeds with an MLP-based primary network). Time measured on Apple M3 Pro (CPU).}
\label{tab:monotone-siren}
\resizebox{\columnwidth}{!}{%
\begin{tabular}{lllllll}
\toprule
Method & $\delta_\text{mae}$ & $\eta_\text{rate}$ & $\eta_\text{mean}$ & $\eta_\text{max}$ & $n_\text{sat}/N$ & Time (s) \\
\midrule
Baseline & 1.0e-06 ± 2.3e-06 & 3.7e-01 ± 1.4e-02 & 3.0e+00 ± 5.8e-01 & 4.3e+01 ± 2.0e+01 & 0/10 & 7.3 \\
Penalty & 6.8e-02 ± 6.3e-03 & 1.6e-02 ± 6.8e-03 & 1.9e-03 ± 2.0e-03 & 3.8e-01 ± 3.4e-01 & 0/10 & 17.4 \\
Lagrangian & 3.3e+00 ± 7.8e-02 & 4.9e-01 ± 3.0e-03 & 8.1e+01 ± 4.7e+01 & 1.0e+03 ± 6.5e+02 & 0/10 & 48.3 \\
Aug. Lag. & 2.0e+00 ± 1.7e+00 & 3.0e-01 ± 2.5e-01 & 2.0e+00 ± 1.8e+00 & 2.3e+01 ± 2.1e+01 & 1/10 & 53.3 \\
N. Slack Var. & 7.1e-02 ± 5.5e-03 & 0 & 0 & 0 & 10/10 & 69.6 \\
CMNN & 8.1e-02 ± 5.7e-03 & 0 & 0 & 0 & 10/10 & 23.1 \\
\bottomrule
\end{tabular}
}
\end{table}

The qualitative picture from the MLP $f^\theta$ table carries over: aside from CMNN, the neural slack variable is the only run to retain feasibility at a competitive fit; the penalty method leaves a positive violation rate; the dense Lagrangian fails to fit the data; the augmented Lagrangian is bimodal across seeds, with high variance in both fit and feasibility.

\begin{figure}[p]
    \centering
    \includegraphics[width=\textwidth]{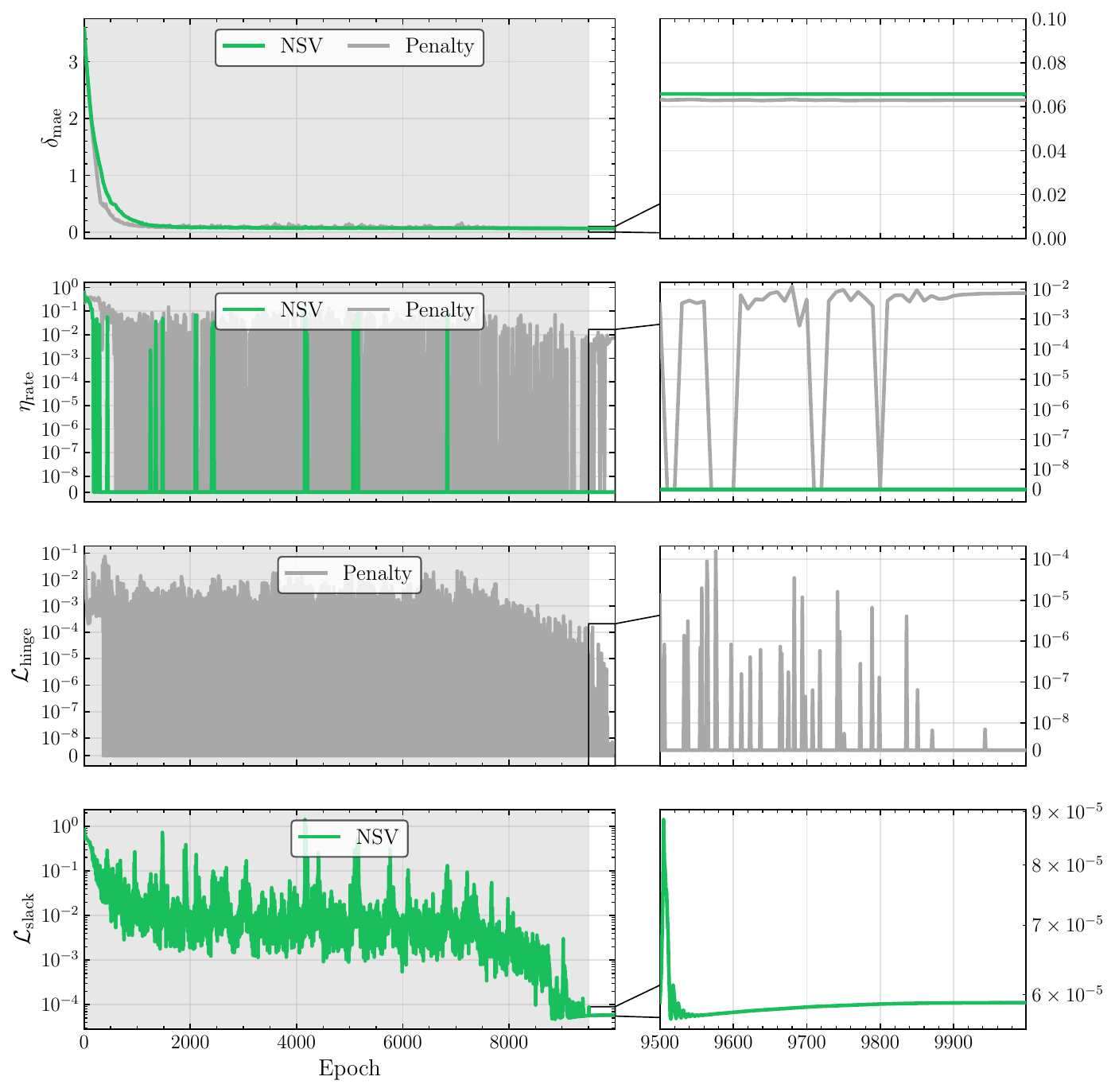}
    \caption{Training dynamics over ten thousand epochs for the monotonicity experiment with a SIREN primary model, with zoom insets (last $500$ epochs) on the right.
    For the representative seed shown, the sampled violation loss becomes small while dense-grid violations persist, whereas the neural slack variable maintains zero dense-grid violations while its matching loss remains active.}
    \label{fig:monotone-siren-training-dynamics}
\end{figure}
\FloatBarrier

\subsection{Neural slack variable capacity sweep}
\label{app:monotone-aux-width}

\begin{figure}[p]
    \centering
    \newcommand{\auxrow}[1]{%
        \begin{subfigure}[t]{\textwidth}
            \includegraphics[width=\textwidth]{resources_v1/monotone/aux_net_ablation_dcr_rel_mlp_width/comparisons/width_#1__seed_42_42.pdf}
            \subcaption{$W = #1$.}
        \end{subfigure}%
    }

    \auxrow{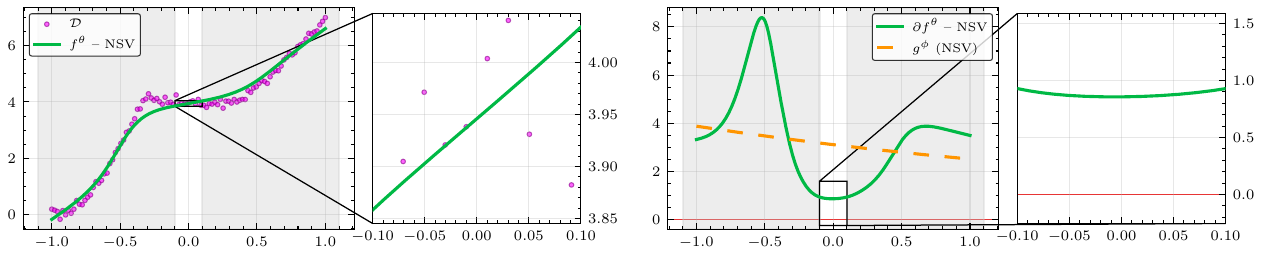}\\[0.5em]
    \auxrow{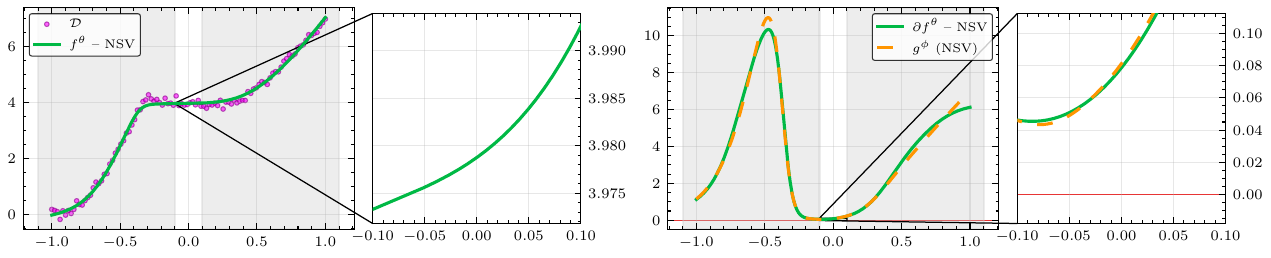}\\[0.5em]
    \auxrow{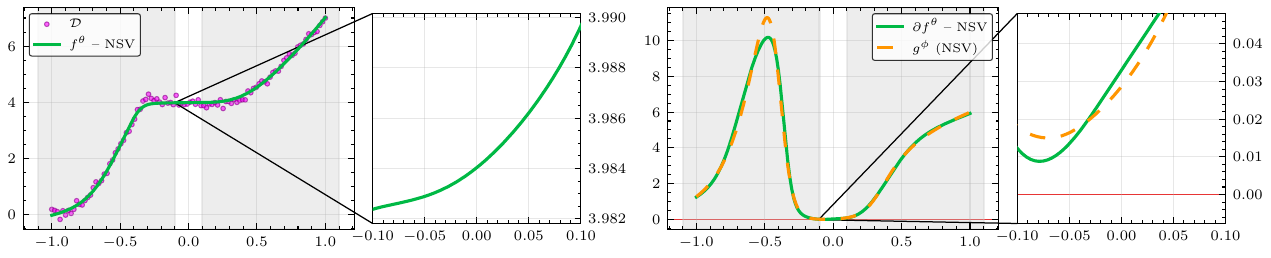}\\[0.5em]
    \auxrow{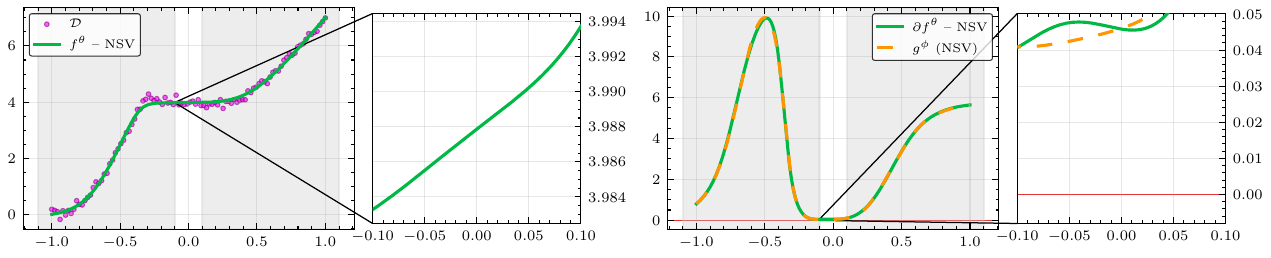}\\[0.5em]
    \auxrow{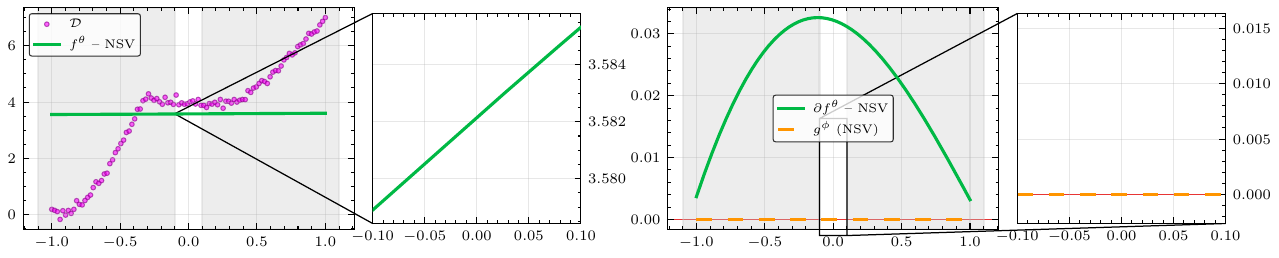}

    \caption{Monotone MLP $+$ neural slack variable, hidden width sweep. Each row pairs the final function fit (left half) with the final derivative profile (right half) at hidden width $W \in \{1, 4, 16, 64, 256\}$ (seed $42$). Shaded grey bands on the full panels mark the zoom region $[-0.1, 0.1]$ rendered alongside.}
    \label{fig:monotone-aux-derivatives}
\end{figure}

Complementing \Cref{sec:exp-monotone-capacity}, \Cref{fig:monotone-aux-derivatives} shows the final $f^\theta$ (left column) and $\partial_x f^\theta$ (right column) for an MLP slack variable at every width in the sweep $W \in \{1, 4, 16, 64, 256\}$, on the monotone MLP setup with seed~$42$.
The two columns track the U-shape visible in \Cref{fig:monotone-aux-capacity-sweep}.
At small widths ($W = 1$) the slack variable has so little capacity that it can only represent a near-constant slack: the constraint pressure pulls $\partial_x f^\theta$ down to a smooth proxy and the plateau is rounded off into a near-linear ramp, costing data fit.
At the optimum ($W = 4$, $16$) the slack tracks the constraint slack accurately and $f^\theta$ resolves both the outer slopes and the central plateau.
At large widths ($W = 64$, $256$) the optimization collapses on this seed: $f^\theta$ degenerates toward a constant predictor and $\partial_x f^\theta$ vanishes almost everywhere on $\Omega$, so the constraint is trivially satisfied at the cost of any data fit.
The collapse is clearly visible in the per-row plots and is the source of the rising $\delta_\text{mae}$ tail in \Cref{fig:monotone-aux-capacity-sweep}.

\begin{figure}[p]
    \centering
    \newcommand{\auxrowsiren}[1]{%
        \begin{subfigure}[t]{\textwidth}
            \includegraphics[width=\textwidth]{resources_v1/monotone/aux_net_ablation_dcr_arch_width_mlp/comparisons/arch_siren__width_#1__seed_42_42.pdf}
            \subcaption{$W = #1$.}
        \end{subfigure}%
    }

    \auxrowsiren{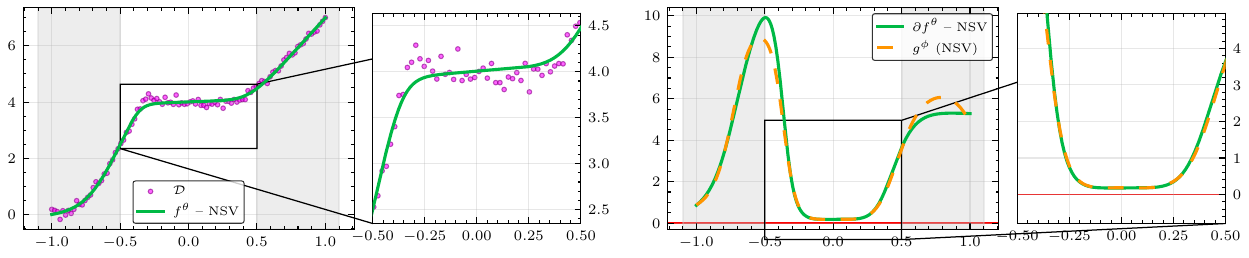}\\[0.5em]
    \auxrowsiren{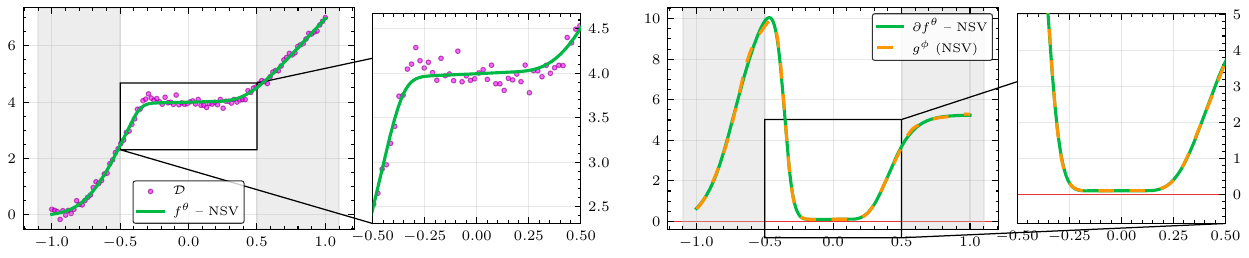}\\[0.5em]
    \auxrowsiren{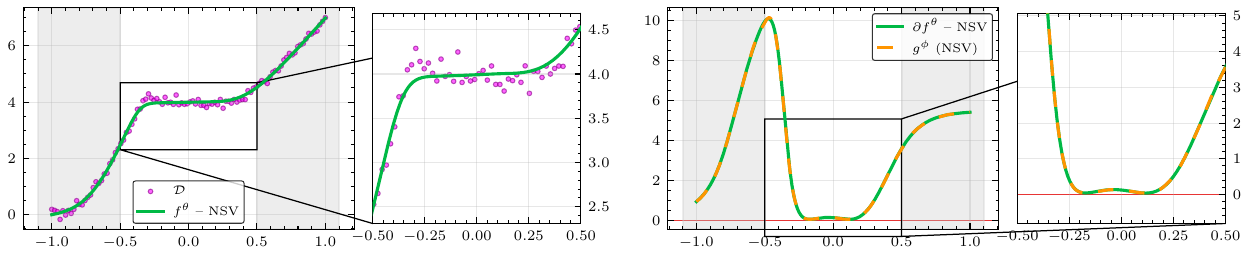}\\[0.5em]
    \auxrowsiren{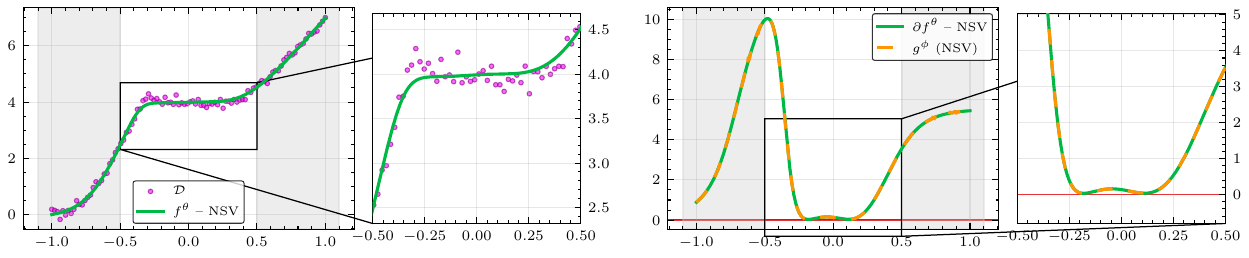}\\[0.5em]
    \auxrowsiren{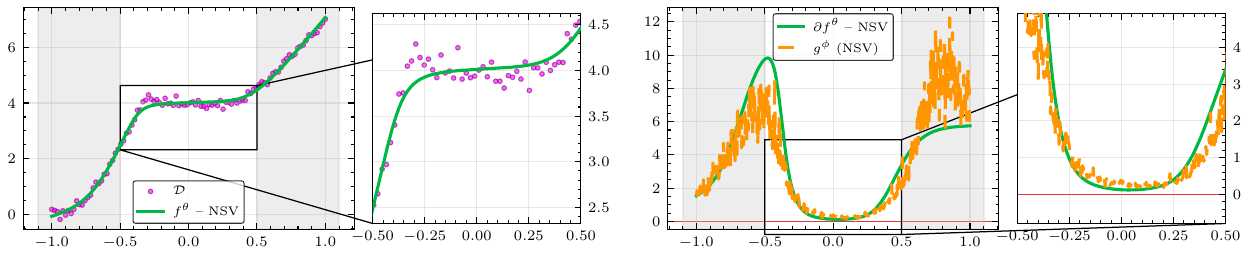}

    \caption{Monotone MLP $+$ SIREN slack variable, hidden width sweep. Counterpart to \Cref{fig:monotone-aux-derivatives} with $s^\phi$ a SIREN ($\omega_0 = 5$) instead of a softplus MLP. Shaded grey bands on the full panels mark the zoom region $[-0.5, 0.5]$ rendered alongside.}
    \label{fig:monotone-aux-derivatives-siren}
\end{figure}

The SIREN slack variable exhibits no analogous U-shape across the sweep: \Cref{fig:monotone-aux-derivatives-siren} renders the per-width fits and derivatives for a SIREN slack on the same MLP $f^\theta$ setup, seed~$42$.
The slack tracks the constraint profile at every $W$, $f^\theta$ retains both the outer slopes and the central plateau, and the collapse modes seen at large $W$ for the MLP slack do not appear.

\section{\texorpdfstring{Additional details: Convexity ($d \in \{2, 3, 4, 5\}$)}{Additional details: Convexity (d in {2, 3, 4, 5})}}
\label{app:convex-nd}

\subsection{Dataset and target function}
\label{app:synthetic-data-convex-nd}

The target is the log-sum-exp of $K = 10$ affine functions plus a small quadratic anchor:
\begin{equation*}
    f(x) = \tau \log \!\left( \sum_{k=1}^K \exp \!\left( \frac{a_k^\top x + b_k}{\tau} \right) \right) + \frac{\mu}{2} \|x\|^2_2, \qquad x \in \Omega = [-1, 1]^d.
\end{equation*}
The affine coefficients $a_k \in \mathbb{R}^d$ and offsets $b_k \in \mathbb{R}$ are drawn i.i.d.\ from $\mathcal{N}(0, I_d)$ and $\mathcal{N}(0, 1)$ at a fixed target seed; the log-sum-exp temperature is $\tau = 0.2$ and the quadratic anchor is $\mu = 10^{-4}$.
Training observations are noisy samples drawn as a scrambled Sobol low-discrepancy sequence on $\Omega$ (additive zero-mean Gaussian noise, standard deviation $0.05$); constraint enforcement uses a separate scrambled Sobol grid (with an optional small absolute pad past $\Omega$); evaluation uses a denser Sobol Monte Carlo sample.
Each $d$ is run with five seeds; the initialization, noise, training-grid, and target seeds advance jointly across the five repetitions, so the same seed index gives a reproducible $({a_k}, {b_k}, \text{init}, \text{noise}, \text{grid})$ tuple.

\subsection{Architectural baseline (ICNN)}
\label{app:convex-nd-icnn}

As an architectural reference we use Input Convex Neural Networks~\cite{amosInputConvexNeural2017} with three hidden layers of width $128$ and $\operatorname{Softplus}$ activations; the output is linear.
The ICNN trains under the same $10{,}000$-epoch Adam schedule as the other comparison runs.

\subsection{Per-dimension tables}
\label{app:convex-nd-tables}

\Cref{tab:convex-nd-summary} reports final metrics at each $d$ as mean $\pm$ std across the five seeds, alongside the unconstrained reference (``None'').
Violation metrics ($\eta_\text{rate}$, $\eta_\text{mean}$, $\eta_\text{max}$, $n_\text{sat}/N$) are defined in \Cref{app:evaluation}; here the constraint components are the Hessian eigenvalues.

\begin{table}[h]
\centering
\caption{Convex-$d$, $d \in \{2, 3, 4, 5\}$ (mean $\pm$ std over 5 seeds). Apple M3 Pro (CPU); total wall-clock $\sim\!20$ h, of which the unconstrained baseline and ICNN training are negligible.}
\label{tab:convex-nd-summary}
\resizebox{\columnwidth}{!}{\begin{tabular}{lllllll}
\toprule
 &  & $\delta_\text{mae}$ & $\eta_\text{rate}$ & $\eta_\text{mean}$ & $\eta_\text{max}$ & $n_\text{sat}/N$ \\
Dim & Method &  &  &  &  &  \\
\midrule
\multirow[t]{4}{*}{2} & None & 4.1e-02 ± 6.9e-04 & 3.8e-01 ± 1.3e-01 & 3.4e-02 ± 4.0e-02 & 9.8e-01 ± 1.6e+00 & 0/5 \\
 & Penalty & 4.1e-02 ± 6.4e-04 & 3.3e-04 ± 6.8e-05 & 4.3e-08 ± 2.4e-08 & 2.2e-03 ± 1.7e-03 & 0/5 \\
 & NSV & 4.2e-02 ± 1.4e-03 & 0 & 0 & 0 & 5/5 \\
 & ICNN & 5.4e-02 ± 6.1e-03 & 0 & 0 & 0 & 5/5 \\
\cline{1-7}
\multirow[t]{4}{*}{3} & None & 4.1e-02 ± 1.2e-03 & 8.0e-01 ± 1.3e-01 & 5.0e-02 ± 3.5e-02 & 1.1e+00 ± 6.8e-01 & 0/5 \\
 & Penalty & 4.3e-02 ± 1.7e-03 & 1.3e-03 ± 6.8e-04 & 1.2e-06 ± 1.7e-06 & 2.1e-02 ± 2.5e-02 & 0/5 \\
 & NSV & 4.4e-02 ± 2.9e-03 & 3.0e-07 ± 5.2e-07 & 1.4e-10 ± 2.9e-10 & 1.0e-03 ± 1.8e-03 & 3/5 \\
 & ICNN & 6.5e-02 ± 1.4e-02 & 0 & 0 & 0 & 5/5 \\
\cline{1-7}
\multirow[t]{4}{*}{4} & None & 4.1e-02 ± 1.1e-03 & 9.2e-01 ± 6.7e-02 & 6.8e-02 ± 2.0e-02 & 2.3e+00 ± 3.4e-01 & 0/5 \\
 & Penalty & 4.7e-02 ± 1.9e-03 & 2.9e-03 ± 1.7e-03 & 4.8e-06 ± 4.9e-06 & 2.7e-01 ± 4.4e-01 & 0/5 \\
 & NSV & 4.8e-02 ± 2.8e-03 & 1.8e-05 ± 1.0e-05 & 3.8e-08 ± 3.6e-08 & 5.8e-02 ± 4.7e-02 & 0/5 \\
 & ICNN & 7.8e-02 ± 1.9e-02 & 0 & 0 & 0 & 5/5 \\
\cline{1-7}
\multirow[t]{4}{*}{5} & None & 4.4e-02 ± 1.2e-03 & 9.6e-01 ± 1.4e-02 & 1.1e-01 ± 1.9e-02 & 5.9e+00 ± 2.4e+00 & 0/5 \\
 & Penalty & 5.9e-02 ± 6.1e-03 & 6.0e-03 ± 1.7e-03 & 1.3e-05 ± 8.0e-06 & 2.6e-01 ± 1.1e-01 & 0/5 \\
 & NSV & 6.2e-02 ± 8.3e-03 & 9.9e-05 ± 8.3e-05 & 1.8e-07 ± 2.1e-07 & 7.5e-02 ± 3.2e-02 & 0/5 \\
 & ICNN & 1.1e-01 ± 1.0e-02 & 0 & 0 & 0 & 5/5 \\
\cline{1-7}
\bottomrule
\end{tabular}
}
\end{table}

\FloatBarrier

\pagebreak

\subsection{Neural slack variable surfaces (\texorpdfstring{$d=2$}{d=2})}
\label{app:convex-nd-d2-surfaces}

\begin{figure}[h]
\centering
\includegraphics[width=0.7\columnwidth]{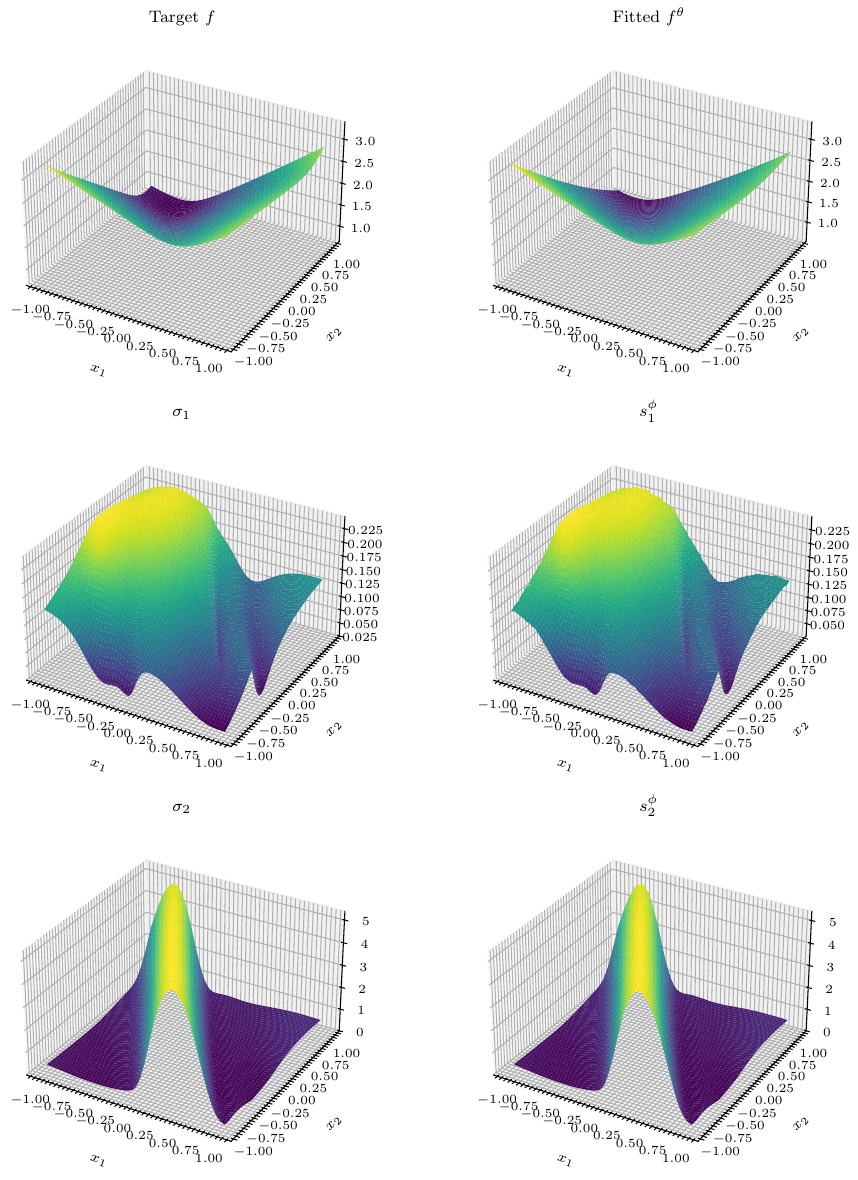}
\caption{Dimension $d = 2$: target $f$, fitted $f^\theta$, sorted Hessian eigenvalues $\sigma_1 \leq \sigma_2$, and neural slack variable outputs $s^\phi_1, s^\phi_2$.}
\label{fig:convex-nd-d2-surfaces}
\end{figure}

\Cref{fig:convex-nd-d2-surfaces} visualizes the target $f$, the fitted $f^\theta$, the actual eigenvalues of $\nabla^2 f^\theta$, and the per-eigenvalue neural slack variable outputs $s^\phi_1, s^\phi_2$ at $d = 2$ for the neural slack variable (rel) run.
The slack outputs are non-negative by construction; the actual eigenvalues track them across $\Omega$.

\FloatBarrier

\section{Additional details: neural barrier certificates}
\label{app:barrier}

\subsection{Domain and constraint geometry}
\label{app:barrier-setup}

The benchmark is the autonomous polynomial system from FOSSIL's \texttt{Barr3} suite~\citep{edwardsFossil20Formal2024},\footnote{\url{https://github.com/oxford-oxcav/fossil}} $\dot x_1 = x_2$, $\dot x_2 = -x_1 - x_2 + \tfrac{1}{3} x_1^3$, on $\Omega = [-3, 2.5] \times [-2, 1]$.
The initial set $\Omega_\mathrm{i}$ and unsafe set $\Omega_\mathrm{u}$ are unions of axis-aligned rectangles and disks, lifted verbatim from the FOSSIL reference configuration:
\begin{align*}
\Omega_\mathrm{i} &= B\!\bigl((1.5, 0),\,0.5\bigr) \;\cup\; [-1.8, -1.2]\!\times\![-0.1, 0.1] \;\cup\; [-1.4, -1.2]\!\times\![-0.5, 0.1], \\
\Omega_\mathrm{u} &= B\!\bigl((-1, -1),\,0.4\bigr) \;\cup\; [0.4, 0.6]\!\times\![0.1, 0.5] \;\cup\; [0.4, 0.8]\!\times\![0.1, 0.3],
\end{align*}
where $B(c, r)$ denotes the closed Euclidean disk.
Boundary samples are drawn by Sobol rejection on each component, with $400$ accepted points each (independent Sobol streams for $\Omega_\mathrm{i}$ and $\Omega_\mathrm{u}$).
The Nagumo training points are $1{,}000$ scrambled Sobol samples on $\Omega$, padded outward by $0.025$ on each axis to soften boundary effects (evaluation uses the unpadded $\Omega$); points are resampled at each training step.

\subsection{Margins, belt option, and Nagumo decay}
\label{app:barrier-knobs}

The strict-positivity margins $\mu_\mathrm{i} = \mu_\mathrm{u} = 10^{-3}$ enter \Cref{eq:barrier-c} as fixed shifts on the boundary components, so the enforced inequalities are $h^\theta \geq \mu_\mathrm{i}$ on $\Omega_\mathrm{i}$ and $h^\theta \leq -\mu_\mathrm{u}$ on $\Omega_\mathrm{u}$.
Margins are distinct from the loss-side $\varepsilon = 10^{-6}$ used inside the violation and slack losses for numerical conditioning.
The Nagumo decay rate is $\gamma = 1$ throughout.

FOSSIL's Barr3 configuration enforces the Nagumo row only on a \emph{belt} around $\{h^\theta = 0\}$, the relaxation justified by Nagumo's theorem requiring $\dot h \geq 0$ only on the boundary of the safe set~\citep{amesControlBarrierFunctions2019}.
We report the unrestricted variant as the default since it provides a stricter operational requirement and a cleaner comparison across methods.

\subsection{Slack-network width sweep}
\label{app:barrier-results}

\Cref{tab:barrier-aux-capacity-siren,tab:barrier-aux-capacity-tanh} report the seed-aggregated metrics behind \Cref{fig:barrier-aux-capacity}: feasibility margin $c^\theta_{\min}$, slack loss $\mathcal{L}_\text{slack}$ at the final epoch, late-stage oscillation $\sigma_\tau$ (std of $\mathcal{L}_\text{slack}$ over the last $1{,}500$ epochs), parameter and output displacements relative to initialization, and the count $n_\text{sat}$ of seeds with $c^\theta_{\min} \geq 0$. Bold $n_\text{sat}/N$ entries flag widths feasible on every seed.

\begin{table}[h]
\centering
\caption{Slack-network width sweep, SIREN slack ($\omega_0 = 5$): per-width seed mean $\pm$ std over $5$ seeds on FOSSIL Barr3. Feasibility holds on every seed at intermediate widths $h \in \{8, 16, 32\}$, peaking at $h = 16$; at the swept extremes the slack collapses to a flat-zero profile (relative output displacement $\approx 1$), and tail oscillation $\sigma_\tau$ traces a U-shape across widths. Apple M3 Pro (CPU); $\sim\!2$ h per sweep.}
\label{tab:barrier-aux-capacity-siren}
\resizebox{\columnwidth}{!}{%
\begin{tabular}{rcccccc}
\toprule
$h$ & $c^\theta_{\min}$ & $\mathcal{L}_\text{slack}$ & $\sigma_\tau$ & $\|\Delta\phi\|_2/\|\phi_0\|_2$ & $\|\Delta s^\phi\|_2/\|s^\phi_0\|_2$ & $n_\text{sat}/N$ \\
\midrule
  1 & -1.0e-02 ± 9.7e-03 & 7.0e-01 ± 1.6e-01 & 3.6e-02 ± 5.1e-02 & 2.5e+01 ± 2.0e+01 & 9.7e-01 ± 2.8e-02 & 0/5 \\
  2 & -9.3e-03 ± 4.2e-03 & 3.4e-01 ± 9.9e-02 & 4.4e-02 ± 5.3e-02 & 9.4e+00 ± 1.5e+00 & 9.0e-01 ± 6.5e-02 & 0/5 \\
  4 & -1.1e-03 ± 5.6e-04 & 1.2e-01 ± 3.5e-02 & 1.0e-02 ± 7.6e-03 & 4.9e+00 ± 6.6e-01 & 1.1e+00 ± 2.5e-01 & 0/5 \\
  8 & +2.4e-03 ± 2.2e-03 & 2.4e-02 ± 1.3e-02 & 4.0e-03 ± 1.2e-03 & 3.3e+00 ± 3.6e-01 & 1.8e+00 ± 6.7e-01 & \textbf{5/5} \\
  16 & +8.7e-03 ± 3.3e-03 & 7.4e-04 ± 2.2e-04 & 7.8e-04 ± 6.0e-04 & 1.8e+00 ± 1.5e-01 & 4.8e+00 ± 1.4e+00 & \textbf{5/5} \\
  32 & +5.2e-03 ± 2.0e-03 & 6.8e-05 ± 2.3e-05 & 3.2e-04 ± 2.4e-04 & 9.6e-01 ± 5.8e-02 & 3.5e+00 ± 1.4e+00 & \textbf{5/5} \\
  64 & +2.0e-04 ± 7.9e-04 & 1.6e-04 ± 1.5e-04 & 1.1e-02 ± 8.3e-03 & 7.4e-01 ± 6.6e-02 & 9.8e-01 ± 1.0e-01 & 3/5 \\
  128 & -6.7e-04 ± 2.1e-04 & 5.1e-04 ± 3.2e-04 & 1.2e-02 ± 1.0e-02 & 6.4e-01 ± 6.7e-02 & 9.4e-01 ± 4.4e-02 & 0/5 \\
  256 & -1.6e-03 ± 5.1e-04 & 1.3e-02 ± 1.4e-02 & 1.3e-01 ± 2.0e-01 & 1.2e+00 ± 3.8e-01 & 9.8e-01 ± 3.3e-02 & 0/5 \\
\bottomrule
\end{tabular}
}
\end{table}

\begin{table}[h]
\centering
\caption{Slack-network width sweep, $\tanh$-MLP slack: per-width seed mean $\pm$ std over $5$ seeds on FOSSIL Barr3. The slack collapses to a flat-zero profile at every swept width (relative output displacement $\approx 1$), failing to inflate. Apple M3 Pro (CPU); $\sim\!2$ h per sweep.}
\label{tab:barrier-aux-capacity-tanh}
\resizebox{\columnwidth}{!}{%
\begin{tabular}{rcccccc}
\toprule
$h$ & $c^\theta_{\min}$ & $\mathcal{L}_\text{slack}$ & $\sigma_\tau$ & $\|\Delta\phi\|_2/\|\phi_0\|_2$ & $\|\Delta s^\phi\|_2/\|s^\phi_0\|_2$ & $n_\text{sat}/N$ \\
\midrule
  1 & -2.6e-03 ± 8.9e-04 & 6.1e-02 ± 4.4e-02 & 1.0e-01 ± 1.4e-01 & 2.4e+00 ± 3.5e-01 & 1.0e+00 ± 2.1e-03 & 0/5 \\
  2 & -2.1e-03 ± 8.1e-04 & 3.0e-02 ± 2.5e-02 & 7.3e-02 ± 1.1e-01 & 2.2e+00 ± 2.4e-01 & 1.0e+00 ± 2.1e-03 & 0/5 \\
  4 & -1.5e-03 ± 1.4e-04 & 1.2e-02 ± 2.7e-03 & 2.7e-02 ± 3.3e-02 & 1.8e+00 ± 1.1e-01 & 1.0e+00 ± 1.2e-03 & 0/5 \\
  8 & -1.4e-03 ± 5.3e-05 & 9.0e-03 ± 5.4e-04 & 1.7e-02 ± 2.4e-02 & 1.4e+00 ± 8.9e-02 & 1.0e+00 ± 1.6e-03 & 0/5 \\
  16 & -1.3e-03 ± 2.6e-05 & 7.8e-03 ± 4.9e-04 & 8.5e-03 ± 1.0e-02 & 1.0e+00 ± 5.1e-02 & 1.0e+00 ± 1.0e-03 & 0/5 \\
  32 & -1.3e-03 ± 3.5e-05 & 7.1e-03 ± 5.6e-04 & 6.8e-03 ± 7.7e-03 & 7.8e-01 ± 4.5e-02 & 1.0e+00 ± 7.0e-04 & 0/5 \\
  64 & -1.3e-03 ± 9.2e-05 & 7.2e-03 ± 1.3e-03 & 3.3e-02 ± 5.8e-02 & 6.0e-01 ± 5.5e-02 & 1.0e+00 ± 1.0e-03 & 0/5 \\
  128 & -1.2e-03 ± 1.3e-04 & 6.2e-03 ± 1.4e-03 & 3.0e-02 ± 5.1e-02 & 4.7e-01 ± 4.0e-02 & 1.0e+00 ± 3.4e-03 & 0/5 \\
  256 & -1.3e-03 ± 1.0e-04 & 7.8e-03 ± 2.1e-03 & 1.7e-02 ± 1.5e-02 & 3.8e-01 ± 5.7e-02 & 1.0e+00 ± 3.2e-03 & 0/5 \\
\bottomrule
\end{tabular}
}
\end{table}

\section{Additional details: Generative volatility surfaces}
\label{app:vol-autodec}

\subsection{Options and implied volatility}
\label{app:vol-autodec-options}

Options are used to hedge against or speculate on price movements of an underlying asset, and they trade in record volumes on stocks, indices, currencies, and commodities.
We focus on options on a major equity index (e.g.\ the Nasdaq-100).
At a given instant $t_0$, the collection of traded options on such an underlying gives a dataset $\mathcal{D} = \{((\tau_i, k_i), y_i)\}_{i=1}^N$, where
\begin{itemize}[leftmargin=1.4em,itemsep=0pt]
    \item $\tau_i = T_i - t_0$ is the \emph{time-to-expiration}, the remaining duration until the option's expiry $T_i$;
    \item $k_i = \log(K_i / F_{T_i})$ is the \emph{log-moneyness}, with strike $K_i$ and forward $F_{T_i}$;\footnote{The forward $F_{T_i}$ is the price of the underlying stripped of interest-rate and dividend effects, projected to expiry $T_i$.}
    \item $y_i$ is the observed \emph{implied volatility}, the standard quoting convention for option prices.
\end{itemize}
\emph{Implied volatility smoothing} is the task of nowcasting a continuous, positive surface $(\tau, k) \mapsto f(\tau, k)$ from $\mathcal{D}$ \citep{gatheralArbitragefreeSVIVolatility2014,lucicNormalizingVolatilityTransforms2021}.
The surface must respect a pair of shape constraints encoding financial consistency.
With \emph{total volatility} $v(\tau, k) := f(\tau, k)\sqrt{\tau}$ and \emph{total variance} $w = v^2$, a surface must be
\begin{itemize}[leftmargin=1.4em,itemsep=0pt]
    \item free from \emph{calendar arbitrage}: $\mathcal{C}_\text{cal}[f] := \partial_\tau w \ge 0$;
    \item free from \emph{butterfly arbitrage}: $\mathcal{C}_\text{str}[f] \ge 0$, where
    \begin{equation*}
        \mathcal{C}_\text{str}[f] := (1 + d_1(\cdot, v)\,\partial_k v)(1 + d_2(\cdot, v)\,\partial_k v) + v\,\partial_k^2 v,
    \end{equation*}
    and $d_{1,2}(\tau, k, v) := -k/v \pm v/2$.
\end{itemize}
The full constraint is $\mathcal{C}[f] := (\mathcal{C}_\text{cal}[f], \mathcal{C}_\text{str}[f])$.

\paragraph{Local volatility}
A no-arbitrage surface yields a \emph{local volatility} model via Dupire's equation,
\begin{equation}\label{eq:Dupire}
    a^2(\tau, k) = \frac{\mathcal{C}_\text{cal}[f](\tau, k)}{\mathcal{C}_\text{str}[f](\tau, k)},
\end{equation}
the workhorse diffusion model for pricing exotic derivatives consistently with observed European option prices.
While no-arbitrage ensures the quotient is well defined, the local volatility inherits its ill-conditioning, often producing irregular surfaces unsuitable for pricing — small residual violations of $\mathcal{C}$ at the level of the implied surface translate into locally undefined or negative local volatility, which makes exact constraint satisfaction crucial in this application.

\subsection{Data}
\label{app:vol-autodec-data}

\paragraph{Source}
We use NDX (Nasdaq-100) option snapshots from OptionMetrics, accessed via WRDS, covering all $N=251$ trading days from 2024-08-29 to 2025-08-29.
Each daily snapshot carries roughly $2{,}300$ mid-quote observations spread across $13$--$18$ listed expiries, with time-to-expiration extending up to $\sim 4.6$ years and log-moneyness spanning roughly $[-1.7, 0.3]$ (NDX puts are listed further from the forward than calls).

\paragraph{Processing}
We average bid and ask quotes for the underlying and the options to obtain mid prices.
Forward prices and discount factors are inferred per expiry from box-spread pricing, with put-call parity as a cross-check.
Time-to-maturity is expressed in year fractions and log-moneyness is computed as in \Cref{app:vol-autodec-options}.
Implied volatilities are obtained from the \emph{py-vollib-vectorized} library, a vectorized implementation of the ``Let's-be-rational'' algorithm of \citet{jackelLetsBeRational2015}.\footnote{\url{https://pypi.org/project/py-vollib-vectorized/}}
We retain out-of-the-money quotes (Put for log-moneyness $k \le 0$, Call for $k > 0$) and discard the rest.
The end-to-end preparation pipeline ships with the supplemental code.

\paragraph{Constraint domain}
$\mathcal{C}[f]$ is a property of the decoded surface and must hold at any latent code, not only at observed $(\tau, k)$, so the constraint domain is decoupled from any per-surface empirical hull.
We enforce constraints on a fixed rectangular box $[\tau_\text{min}, \tau_\text{max}] \times [k_\text{min}, k_\text{max}] = [0.01, 1.0] \times [-0.5, 0.5]$.
The training data are restricted to the same $\tau$ range (the closest expiry slice on each side is retained for boundary stability), with no $k$ filter applied; this leaves roughly $2{,}000$ observations across $12$--$13$ expiries per surface.

\subsection{Autodecoder model}
\label{app:vol-autodec-model}

Each surface receives a learned latent code $z_i \in \mathbb{R}^{32}$, optimized jointly with the decoder by gradient descent (no encoder, following the variational-autodecoder formulation of \citealt{zadehVariationalAutoDecoderMethod2021}).
The decoder maps $(z_i, \tau, k)$ to implied volatility, with moneyness rescaled by $\sqrt{\tau}$ to exploit smile-shape stationarity across expiries and $\tau$ encoded via a log-spaced Fourier embedding.
Its backbone is a learnable tanh feature layer followed by a two-hidden-layer SIREN tail.
The tanh layer can be read as a low-frequency ($\omega_0 \approx 1$) SIREN layer that saturates rather than oscillating outside the active region, giving graceful asymptotics at large absolute log-moneyness; the higher-frequency SIREN tail then represents fine smile structure.
The latent code and the time embedding are concatenated into a per-slice conditioning vector for a small FiLM modulator that produces $(\gamma_\ell, \beta_\ell)$ for every backbone layer, including the tanh layer; $\gamma_\ell$ uses a learnable scaled sigmoid so FiLM can both attenuate and amplify.

\subsection{Training}
\label{app:vol-autodec-training}

$100$ epochs of Adam with separate learning rates for decoder/slack parameters ($3 \times 10^{-4}$) and per-surface latent codes ($10^{-3}$); batch size $8$ surfaces, gradient clipping at norm~$1$.
A KL term (weight $\beta = 10^{-2}$) regularizes each $z_i$ towards $\mathcal{N}(0, I)$, controlling the geometry of the trained aggregate posterior used by the \textsc{prior} evaluation regime below.
The constraint grid is a Sobol cloud of $10{,}000$ $(\tau, k)$ points per step over $[\tau_\text{min}, \tau_\text{max}] \times [k_\text{min}, k_\text{max}]$, with $\tau$ warped through $u \mapsto (\sqrt{\tau_\text{min}} + u(\sqrt{\tau_\text{max}}-\sqrt{\tau_\text{min}}))^2$.
For the penalty method and the augmented Lagrangian, the constraint weight is increased multiplicatively on a tolerance trigger and capped at $100$; under this long-run minibatch regime with a sparse constraint grid, larger caps introduce optimization instabilities, consistent with the practice of \citet{ackererDeepSmoothingImplied2020}, who report $100$ as the largest weight tested.

\subsection{Evaluation regimes}
\label{app:vol-autodec-eval}

We evaluate constraint satisfaction on a $200\times200$ Cartesian $(\tau, k)$ grid for each of three latent populations:
\begin{itemize}[leftmargin=1.4em,itemsep=0pt]
    \item \textsc{train}: the $251$ trained latent codes themselves;
    \item \textsc{interpolation}: $100$ samples $z = \alpha \mu_i + (1 - \alpha) \mu_j$, with $\mu_i, \mu_j$ a uniformly sampled pair of trained codes and $\alpha \sim \mathrm{Uniform}(0, 1)$;
    \item \textsc{prior}: $100$ samples from $\mathcal{N}(\bar{\mu}, \widehat{\Sigma})$, the moment-matched Gaussian approximation to the trained aggregate posterior $q(z) = \tfrac{1}{N}\sum_i q(z \mid x_i)$. Here $\bar{\mu}$ is the empirical mean of the trained $\mu_i$ and $\widehat{\Sigma}$ is their empirical covariance plus the average per-surface posterior variance on the diagonal (the within/between decomposition of $\mathrm{Cov}_{q(z)}[z]$). Sampling from $\mathcal{N}(\bar{\mu}, \widehat{\Sigma})$ rather than the formal $\mathcal{N}(0, I)$ prior accounts for the gap that the small KL weight $\beta = 10^{-2}$ leaves between $q(z)$ and the prior.
\end{itemize}

\subsection{Neural augmented Lagrangian and neural slack variable}
\label{app:vol-autodec-baselines}

The penalty baseline adds the standard $p\!=\!1$ hinge penalty from \eqref{eq:bg-hinge} to the calendar and strike no-arbitrage constraint heads.
The adaptive penalty weight is capped at $100$ in the reported run.

The neural augmented Lagrangian and neural slack variable baselines use the same FiLM conditioning convention as the volatility decoder, but use plain SIREN networks with two output channels and no initial $\tanh$ feature layer.
They take $k/\sqrt{\tau}$ as the scalar network input and use the concatenated vector $[z_i,\Phi(\tau)]$ for FiLM conditioning.

For the neural augmented Lagrangian the two channels are the multipliers for the calendar and strike no-arbitrage constraints.
We train the multiplier network jointly with the autodecoder under the augmented Lagrangian objective, using joint optimization for the primal and multiplier parameters.
The adaptive penalty weight is also capped at $100$.

For the neural slack variable baseline, the two output channels are activated with $\varepsilon+\exp(\cdot)$.
This gives one positive slack variable in $[\varepsilon,\infty)$ for each no-arbitrage constraint.

%\newpage
%\input{checklist.tex}

\end{document}